\documentclass{article}

\PassOptionsToPackage{numbers, compress}{natbib}

\usepackage[final]{neurips_2019}

\usepackage[utf8]{inputenc} 
\usepackage[T1]{fontenc}    
\usepackage{hyperref}       
\usepackage{url}            
\usepackage{booktabs}       
\usepackage{amsfonts}       
\usepackage{nicefrac}       
\usepackage{microtype}      
\usepackage[dvipsnames]{xcolor}
\usepackage{amsmath}
\usepackage{amssymb}
\usepackage{amsthm}
\usepackage{graphicx}
\usepackage{dsfont}
\usepackage{tikz}
\usepackage{subcaption}
\usepackage{array}
\usepackage{makecell}
\usepackage{multirow}
\usepackage{wrapfig}
\usepackage{wasysym}
\usetikzlibrary{positioning}

\newtheorem{proposition}{Proposition}
\newtheorem{theorem}{Theorem}
\newtheorem{lemma}{Lemma}
\newtheorem{corollary}{Corollary}

\newtheorem{definition}{Definition}
\newtheorem*{theorem*}{Theorem}

\newcommand{\R}{\mathbb{R}}
\newcommand{\E}{\mathop{\mathbb{E}}}
\newcommand{\Z}{\mathcal{Z}}
\newcommand{\X}{\mathcal{X}}

\newcommand{\V}{\text{Var}}

\newcommand{\XN}{\mathbf{X}^N}
\newcommand{\XM}{\mathbf{X}^M}
\newcommand{\ZM}{\mathbf{Z}^M}

\newcommand{\KL}{\mathrm{KL}}
\newcommand{\TV}{\mathrm{TV}}
\newcommand{\JS}{\mathrm{JS}}
\newcommand{\Hsq}{\mathrm{H}^2}

\newcommand\encircle[1]{%
  \tikz[baseline=(X.base)] 
    \node (X) [draw, shape=circle, inner sep=0] {\strut #1};}

\definecolor{transparent}{RGB}{200, 200, 200}
\definecolor{darkgreen}{RGB}{0, 170, 0}
\definecolor{lightgreen}{RGB}{0, 200, 0}
\definecolor{ineq}{RGB}{220, 220, 255}
\definecolor{purple}{RGB}{153, 0, 255}
\definecolor{amber}{rgb}{1.0, 0.75, 0.0}

\title{Practical and Consistent Estimation of $f$-Divergences}

\author{%
  Paul K. Rubenstein\thanks{Part of this work was done during an internship at Google.} \\
  Max Planck Institute for Intelligent Systems, T\"ubingen \\
  \& Machine Learning Group, University of Cambridge \\
  \texttt{paul.rubenstein@tuebingen.mpg.de} \\
  \And
  Olivier Bousquet, Josip Djolonga, Carlos Riquelme, Ilya Tolstikhin \\
  Google Research, Brain Team, Z\"urich \\
  \texttt{\{obousquet, josipd, rikel, tolstikhin\}@google.com}
}

\begin{document}

\maketitle

\begin{abstract}
The estimation of an $f$-divergence between two probability distributions based on samples is a fundamental problem in statistics and machine learning.
Most works study this problem under very weak assumptions, in which case it is provably hard.
We consider the case of stronger structural assumptions that are commonly satisfied in modern machine learning, including  representation learning and generative modelling with autoencoder architectures.
Under these assumptions we propose and study an estimator that can be easily implemented, works well in high dimensions, and enjoys faster rates of convergence.
We verify the behavior of our estimator empirically in both synthetic and real-data experiments, and discuss its direct implications for total correlation, entropy, and mutual information estimation.
\end{abstract}
\section{Introduction and related literature}\label{sec:intro}

The estimation and minimization of divergences between probability distributions based on samples are fundamental problems of machine learning.
For example, maximum likelihood learning can be viewed as minimizing the Kullback-Leibler divergence $\KL(P_{\text{data}}\| P_{\text{model}})$ with respect to the model parameters. 
More generally, generative modelling---most famously Variational Autoencoders and Generative Adversarial Networks \cite{kingma2013auto, goodfellow2014generative}---can be viewed as minimizing a divergence $\smash{D(P_{\text{data}}\| P_{\text{model}})}$ where $\smash{P_{\text{model}}}$ may be intractable.
In variational inference, an intractable posterior $p(z|x)$ is approximated with a tractable distribution $q(z)$ chosen to minimize $\smash{\KL\bigl(q(z) \| p(z|x)\bigr)}$.
The mutual information between two variables $\smash{I(X,Y)}$, core to information theory and Bayesian machine learning, is equivalent to $\smash{\KL(P_{X,Y} \| P_X P_Y)}$. 
Independence testing often involves estimating a divergence $\smash{D(P_{X,Y} \| P_X P_Y)}$, while two-sample testing (does $P=Q$?) involves estimating a divergence $D(P\|Q)$.
Additionally, one approach to domain adaptation, in which a classifier is learned on a distribution $P$ but tested on a distinct distribution $Q$, involves learning a feature map $\phi$ such that a divergence $\smash{D\left( \phi_\# P \| \phi_\# Q \right)}$ is minimized, where $\smash{\phi_\#}$ represents the push-forward operation \cite{ben2007analysis,ganin2016domain}.

In this work we consider the well-known family of $f$-divergences \cite{csiszar2004information, liese2006divergences} that includes amongst others the $\KL$, Jensen-Shannon ($\JS$), $\chi^2$, and $\alpha$-divergences as well as the Total Variation ($\TV$) and squared Hellinger ($\Hsq$) distances, the latter two of which play an important role in the statistics literature \cite{tsybakov2009}.
A significant body of work exists studying the estimation of the $f$-divergence $D_f(Q \| P)$ between general probability distributions $Q$ and $P$.
While the majority of this focuses on $\alpha$-divergences and closely related R\'enyi-$\alpha$ divergences \citep{poczos11alpha, singh14alpha, krishnamurthy14icml},
many works address specifically the KL-divergence \citep{perez08kl, wang09kl}
with fewer considering $f$-divergences in full generality \cite{nguyen10ratio, kanamori12ratio, moon14ensemble, moon14followup}.
Although the $\KL$-divergence is the most frequently encountered $f$-divergence in the machine learning literature, in recent years there has been a growing interest in other $f$-divergences \cite{nowozin2016f}, 
in particular in the variational inference community where they have been employed to derive alternative evidence lower bounds \cite{pmlr-v80-chen18k, li2016renyi, dieng2017variational}.

The main challenge in computing $D_f(Q \| P)$ is that it requires knowledge of either the densities of both $Q$ and $P$, or the density ratio $dQ/dP$.
In studying this problem, assumptions of differing strength can be made about $P$ and $Q$. 
In the weakest \emph{agnostic} setting, we may be given only a finite number of i.i.d\:samples from the distributions without any further knowledge about their densities.
As an example of stronger assumptions,
both distributions may be mixtures of Gaussians \cite{hershey2007approximating, durrieu2012lower}, 
or we may have access to samples from $Q$ and have full knowledge of $P$ \citep{heroma2001techrep, heroma2002ieee} as in e.g.\:model fitting.

Most of the literature on $f$-divergence estimation considers the weaker agnostic setting.
The lack of assumptions makes such work widely applicable, but comes at the cost of needing to work around estimation of either the densities of $P$ and $Q$ \cite{singh14alpha, krishnamurthy14icml} or the density ratio $dQ/dP$ \citep{nguyen10ratio, kanamori12ratio} from samples.
Both of these estimation problems are provably hard \citep{tsybakov2009, nguyen10ratio} and suffer rates---the speed at which the error of an estimator decays as a function of the number of samples $N$---of order $\smash{N^{-1/d}}$ when $P$ and $Q$ are defined over $\R^d$ unless their densities are sufficiently smooth.
This is a manifestation of the \emph{curse of dimensionality} and rates of this type are often called \emph{nonparametric}.
One could hope to estimate $D_f(P\|Q)$ without explicitly estimating the densities or their ratio and thus avoid suffering nonparametric rates, however a lower bound of the same order $\smash{N^{-1/d}}$ was recently proved for $\alpha$-divergences \citep{krishnamurthy14icml}, a sub-family of $f$-divergences.
While some works considering the agnostic setting provide rates for the bias and variance of the proposed estimator \cite{nguyen10ratio, krishnamurthy14icml} or even exponential tail bounds \citep{singh14alpha},
it is more common to only show that the estimators are asymptotically unbiased or consistent without proving specific rates of convergence \cite{wang09kl, poczos11alpha, kanamori12ratio}.

Motivated by recent advances in machine learning, we study a setting in which much stronger structural assumptions are made about the distributions.
Let $\X$ and $\Z$ be two finite dimensional Euclidean spaces.
We estimate the divergence $D_f(Q_Z\| P_Z)$ between two probability distributions $P_Z$ and $Q_Z$, both defined over $\Z$.
$P_Z$ has known density $p(z)$, while $Q_Z$ with density  $q(z)$ admits the factorization $\smash{q(z) := \int_\X q(z|x)q(x) dx}$ where access to independent samples from the distribution $Q_X$ with unknown density $q(x)$ and full knowledge of the conditional distribution $\smash{Q_{Z|X}}$ with density $q(z|x)$ are assumed.
In most cases $Q_Z$ is intractable due to the integral and so is $D_f(Q_Z \| P_Z)$.
As a concrete example, these assumptions are often satisfied in applications of modern unsupervised generative modeling with deep autoencoder architectures,
where $\X$ and $\Z$ would be \emph{data} and \emph{latent} spaces, $\smash{P_Z}$ the \emph{prior}, $\smash{Q_X}$ the \emph{data distribution}, $\smash{Q_{Z|X}}$ the \emph{encoder}, and $\smash{Q_Z}$ the \emph{aggregate posterior}.

Given independent observations $\smash{X_1, \ldots, X_N}$ from $\smash{Q_X}$, the finite mixture $\smash{\hat{Q}_Z^N := \frac{1}{N} \sum_{i=1}^N Q_{Z|X_i}}$ can be used to approximate the continuous mixture $\smash{Q_Z}$. 
{\bf Our main contribution} is to approximate the intractable $\smash{D_f(Q_Z \| P_Z)}$ with $\smash{D_f(\hat{Q}_Z^N \| P_Z)}$, a quantity that can be estimated to arbitrary precision using Monte-Carlo sampling since both distributions have known densities, and to theoretically study conditions under which this approximation is reasonable.
We call $\smash{D_f(\hat{Q}_Z^N \| P_Z)}$ the Random Mixture (RAM) estimator and derive rates at which it converges to $\smash{D_f(Q_Z \| P_Z)}$ as $N$ grows.
We also provide similar guarantees for RAM-MC---a practical Monte-Carlo based version of RAM.
By side-stepping the need to perform density estimation, we obtain \emph{parametric} rates of order $N^{-\gamma}$, where $\gamma$ is independent of the dimension (see Tables \ref{table:convergence} and \ref{table:concentration}), although the constants may still in general show exponential dependence on dimension.
This is in contrast to the agnostic setting where \emph{both} nonparametric rates and constants are exponential in dimension. 

Our results have immediate implications to existing literature.
For the particular case of the $\KL$ divergence, a similar approach has been \emph{heuristically} applied independently by several authors for estimating the mutual information \cite{poolevariational} and total correlation \cite{chen2018isolating}.
Our results provide strong theoretical grounding for these existing methods by showing sufficient conditions for their consistency.

A final piece of related work is \cite{burda2015importance}, which proposes to reduce the gap introduced by Jensen's inequality in the derivation of the classical evidence lower bound (ELBO) by using multiple Monte-Carlo samples from the approximate posterior $\smash{Q_{Z|X}}$.
This is similar in flavour to our approach, but fundamentally different since we use multiple samples from the \emph{data distribution} to reduce a different Jensen gap.
To avoid confusion, we note that replacing the ``regularizer'' term $\mathbb{E}_X[\KL(Q_{Z|X} \| P_Z)]$ of the classical ELBO with expectation of our estimator $\E_{\XN}[\KL(\hat{Q}_Z^N\| P_Z)]$ results in an upper bound of the classical ELBO (see Proposition~\ref{prop:upper-bound}) but is itself not in general an evidence lower bound:
{\addtolength{\abovedisplayskip}{-0.0mm}
\addtolength{\belowdisplayskip}{-3.0mm}
\begin{align*}
    \mathbb{E}_X \Big[ \mathbb{E}_{Q_{Z|X}} \log p(X|Z) - \KL(Q_{Z|X} \| P_Z ) \Big] \leq \mathbb{E}_X \Big[ \mathbb{E}_{Q_{Z|X}} \log p(X|Z) \Big] - \mathbb{E}_{\XN} \Big[ \KL(\hat{Q}_Z^N \| P_Z ) \Big].
\end{align*}}

The remainder of the paper is structured as follows.
In Section \ref{sec:theory} we introduce the RAM and RAM-MC estimators and present our main theoretical results, including rates of convergence for the bias (Theorems~\ref{thm:fast-KL-rate} and \ref{thm:convergence-rate-general}) and tail bounds (Theorems \ref{thm:concentration} and \ref{thm:mc-variance}).
In Section \ref{sec:experiments} we validate our results in both synthetic and real-data experiments. 
In Section \ref{sec:applications} we discuss further applications of our results.
We conclude in Section \ref{sec:conclusion}.

\section{Random mixture estimator and convergence
results}\label{sec:theory}
In this section we introduce our $f$-divergence estimator, and present theoretical guarantees for it.
We assume the existence of probability distributions
${P_Z}$ and ${Q_Z}$ defined over $\Z$ with known density $p(z)$ and intractable density ${q(z) = \int q(z|x) q(x) dx}$ respectively,  where ${Q_{Z|X}}$ is known. $Q_X$ defined over $\X$ is unknown, however we have an i.i.d.\:sample ${\XN=\{X_1, \ldots, X_N\}}$ from it.
Our ultimate goal is to estimate the intractable $f$-divergence $D_f(Q_Z \| P_Z)$ defined by:
\begin{definition}[$f$-divergence]
\label{def:fdiv}
Let $f$ be a convex function on $(0, \infty)$ with $f(1) = 0$. 
The $f$-divergence $D_f$ between distributions $Q_Z$ and $P_Z$ admitting densities $q(z)$ and $p(z)$ respectively is
{\addtolength{\abovedisplayskip}{-0.5mm}
\addtolength{\belowdisplayskip}{-0.5mm}
\begin{align*}
    D_f(Q_Z \| P_Z) := \int f \left( \frac{q(z)}{p(z)} \right) p(z) dz.
\end{align*}}%
\end{definition}
Many commonly used divergences such as Kullback–Leibler and $\chi^2$ are $f$-divergences.
All the divergences considered in this paper together with their corresponding $f$ can be found in Appendix~\ref{appendix:f-fns}. 
Of them, possibly the least well-known in the machine learning literature are $f_\beta$-divergences \cite{osterreicher2003new}. 
These symmetric divergences are continuously parameterized by $\beta\in(0, \infty]$. Special cases include squared-Hellinger ($\mathrm{H}^2$) for ${\beta=\frac{1}{2}}$,  Jensen-Shannon (JS) for $\beta=1$, Total Variation (TV) for $\beta=\infty$. 

In our setting $Q_Z$ is intractable and so is ${D_f(Q_Z \| P_Z)}$.
Substituting $Q_Z$ with a sample-based finite mixture ${\hat{Q}_Z^N := \frac{1}{N} \sum_{i=1}^N Q_{Z|X_i}}$ leads to our proposed 
{\bf Random Mixture estimator (RAM)}:
{\addtolength{\abovedisplayskip}{-1mm}
\addtolength{\belowdisplayskip}{-1mm}
\begin{equation}\textstyle
    D_f\bigl(\hat{Q}_Z^N \| P_Z\bigr) := D_f\Big(\frac{1}{N} \sum_{i=1}^N Q_{Z|X_i} \big\| P_Z\Big).  
\end{equation}}%
Although $\smash{\hat{Q}_Z^N}$ is a function of $\smash{\XN}$ we omit this dependence in notation for brevity. 
In this section we identify sufficient conditions under which $\smash{D_f(\hat{Q}_Z^N \| P_Z)}$ is a ``good'' estimator of $\smash{D_f(Q_{Z} \| P_Z)}$.
More formally, we establish conditions under which the estimator is asymptotically unbiased, concentrates to its expected value and can be practically estimated using Monte-Carlo sampling.

\subsection{Convergence rates for the bias of RAM}
The following proposition shows that $D_f(\hat{Q}_Z^N \| P_Z)$ upper bounds $D_f(Q_{Z} \| P_Z)$ in expectation for any finite $N$, and that the upper bound becomes tighter with increasing $N$:
\begin{proposition}\label{prop:upper-bound}
Let $M \leq N$ be integers. Then
\begin{align}
\label{eq:our-estimate}
    D_f(Q_Z \| P_Z) \ \leq 
    \mathbb{E}_{\mathbf{X}^N} \bigl[D_f(\hat{Q}_Z^N \| P_Z)\bigr] \  \leq \ \mathbb{E}_{\mathbf{X}^M} \bigl[D_f(\hat{Q}_Z^M \| P_Z)\bigr].
\end{align}
\end{proposition}
\begin{proof}[Proof sketch (full proof in Appendix \ref{proof:prop1})]
The first inequality follows from Jensen's inequality, using the facts that $f$ is convex and ${Q_Z = \E_{\XN} [\hat{Q}_Z^N}]$.
The second holds since a sample ${\XM}$ can be drawn by sub-sampling (without replacement) $M$ entries of ${\XN}$, and by applying Jensen again.
\end{proof}
As a function of $N$, the expectation is a decreasing sequence that is bounded below.
By the monotone convergence theorem, the sequence converges.
Theorems \ref{thm:fast-KL-rate} and \ref{thm:convergence-rate-general} in this section give sufficient conditions under which the expectation of RAM converges to $D_f(Q_{Z} \| P_Z)$ as $N\to\infty$ for a variety of $f$ and provide rates at which this happens, summarized in Table \ref{table:convergence}.
The two theorems are proved using different techniques and assumptions. 
These assumptions, along with those of existing methods (see Table~\ref{table:convergence-other}) are discussed at the end of this section.

\renewcommand{\arraystretch}{1}
\begin{table}
 \caption{Rate of bias $\E_{\XN} D_f\big(\hat{Q}^N_{Z} \| P_Z\big) - D_f\left(Q_{Z} \| P_Z\right)$.}
 \label{table:convergence}
 \centering
 \begin{tabular}{c c c c c c c c c } 
 \toprule
 \multirow{2}{*}{$f$-divergence} & \multirow{2}{*}{KL} & \multirow{2}{*}{TV} & \multirow{2}{*}{$\chi^2$} & \multirow{2}{*}{$\text{H}^2$} & \multirow{2}{*}{JS} & \multicolumn{2}{c}{\thead{$D_{f_\beta}$}}  & \thead{$D_{f_\alpha}$} \\ [-0.8ex]
 & & & & & & $\scriptstyle{\frac{1}{2}<\beta<1}$ & $\scriptstyle{1<\beta<\infty}$ &
$\scriptstyle{-1<\alpha<1}$ \\
 \midrule
 \thead{Theorem 1} & $\scriptstyle{N^{-1}}$ & $\scriptstyle{N^{-\frac{1}{2}}}$ & - & $\scriptstyle{N^{-\frac{1}{2}}}$ & $\scriptstyle{N^{-\frac{1}{4}}}$ & $\scriptstyle{N^{-\frac{1}{4}}}$ & $\scriptstyle{N^{-\frac{1}{4}}}$ & - \\ 
 \thead{Theorem 2} & $\scriptstyle{N^{-\frac{1}{3}}\log N}$ & $\scriptstyle{N^{-\frac{1}{2}}}$ & $\scriptstyle{N^{-1}}$ & $\scriptstyle{N^{-\frac{1}{5}}}$ & $\scriptstyle{N^{-\frac{1}{3}}\log N}$ & $\scriptstyle{N^{-\frac{1}{3}}}$ & $\scriptstyle{N^{-\frac{1}{2}}}$ & $\scriptstyle{N^{-\frac{\alpha+1}{\alpha+5}}}$ \\
 \bottomrule
\end{tabular}
\end{table}

\begin{theorem}[Rates of the bias]\label{thm:fast-KL-rate}
If
$\E_{X\sim Q_X}\bigl[\chi^2\bigl(Q_{Z|X}, Q_Z\bigr)\bigr]$ and
$\KL\left( Q_{Z} \| P_Z\right)$ are finite then the bias ${\E_{\XN}\bigl[D_f( \hat{Q}_Z^N \| P_Z)\bigr] - D_f\left( Q_{Z} \| P_Z\right)}$ decays with rate as given in the first row of Table~\ref{table:convergence}.
\end{theorem}
\begin{proof}[Proof sketch (full proof in Appendix \ref{appendix:subsec:thm1})]
There are two key steps to the proof. 
The first is to bound the bias by ${\E_{\XN}\big[D_f(\hat{Q}_Z^N, Q_Z)\big]}$. 
For the KL this is an equality. 
For ${D_{f_\beta}}$ this holds because for $\beta {\geq} 1/2$ it is a \emph{Hilbertian metric} and its square root satisfies the triangle inequality \citep{hein05hilbertian}.
The second step is to bound ${\E_{\XN}\bigl[D_f(\hat{Q}_Z^N, Q_Z)\bigr]}$ in terms of ${\E_{\XN}\bigl[\chi^2(\hat{Q}_Z^N, Q_Z)\bigr]}$, which is the variance of the average of $N$ i.i.d.\:random variables and therefore decomposes as ${\E_{X\sim Q_X}\bigl[\chi^2(Q_{Z|X}, Q_Z)\bigr] / N}$.
\end{proof}
\begin{theorem}[Rates of the bias]\label{thm:convergence-rate-general}
If $\E_{X\sim Q_X, Z\sim P_Z}\bigl[ q^4(Z|X) / p^4(Z) \bigr]$ is finite then
the bias $\E_{\XN}\bigl[D_f( \hat{Q}_Z^N \| P_Z)\bigr] - D_f\left( Q_{Z} \| P_Z\right)$ decays with rate as given in the second row of Table \ref{table:convergence}.
\end{theorem}
\begin{proof}[Proof sketch (full proof in Appendix \ref{proof:thm2})]
Denoting by $\hat{q}_N(z)$ the density of $\hat{Q}_Z^N$,
the proof is based on the inequality
$f\bigl(\hat{q}_N(z) / p(z)\bigr) - f\bigl(q(z) / p(z)\bigr)\leq \frac{\hat{q}_N(z) - q(z)}{p(z)} f'\bigl(\hat{q}_N(z) / p(z)\bigr)$ due to convexity of $f$, applied to the bias.
The integral of this inequality is bounded by controlling $f'$, requiring subtle treatment when $f'$ diverges when the density ratio $\hat{q}_N(z)/p(z)$ approaches zero.
\end{proof}

\subsection{Tail bounds for RAM and practical estimation with RAM-MC}

Theorems \ref{thm:fast-KL-rate} and \ref{thm:convergence-rate-general} describe the convergence of the  \emph{expectation} of RAM over $\XN$, which in practice may be intractable.
Fortunately, the following shows that RAM rapidly concentrates to its expectation.

\begin{table}
 \caption{Rate $\psi(N)$ of high probability bounds for $D_f\big(\hat{Q}^N_{Z} \| P_Z\big)$ (Theorem 3).}
 \label{table:concentration}
 \centering
 \begin{tabular}{c c c c c c c c c } 
 \toprule
 \multirow{2}{*}{$f$-divergence} & \multirow{2}{*}{KL} & \multirow{2}{*}{TV} & \multirow{2}{*}{$\chi^2$} & \multirow{2}{*}{$\text{H}^2$} & \multirow{2}{*}{JS} & \multicolumn{2}{c}{\thead{$D_{f_\beta}$}}  & \thead{$D_{f_\alpha}$} \\ [-0.8ex]
 & & & & & & $\scriptstyle{\frac{1}{2}<\beta<1}$ & $\scriptstyle{1<\beta<\infty}$ &
$\scriptstyle{\frac{1}{3}<\alpha<1}$ \\
 \midrule
 \thead{$\psi(N)$} &  $\scriptstyle{N^{-\frac{1}{6}}\log N}$ & $\scriptstyle{N^{-\frac{1}{2}}}$ & 
 $\scriptstyle{N^{-\frac{1}{2}}}$ &
 - & 
 $\scriptstyle{N^{-\frac{1}{6}}\log N}$ &
 $\scriptstyle{N^{-\frac{1}{6}}}$ &
 $\scriptstyle{N^{-\frac{1}{2}}}$ &
 $\scriptstyle{N^{\frac{1-3\alpha}{\alpha+5}}}$
 \\ 
 \bottomrule
\end{tabular}
\end{table}

\begin{theorem}[Tail bounds for RAM]\label{thm:concentration}
Suppose that ${\chi^2\left(Q_{Z|x} \| P_Z\right) \leq C < \infty}$ for all $x$ and for some constant $C$.
Then, the RAM estimator ${D_f( \hat{Q}_Z^N \| P_Z)}$ concentrates to its mean in the following sense. 
For $N>8$ and for any $\delta >0$, with probability at least $1-\delta$ it holds that
\begin{align*}
    \left| D_f( \hat{Q}_Z^N \| P_Z) - \mathbb{E}_{\XN} \bigl[D_f(\hat{Q}_Z^N \| P_Z)\bigr] \right| \leq {K \cdot \psi(N)} \  \sqrt{\log (2/\delta)},
\end{align*}
where $K$ is a constant and $\psi(N)$ is given in Table~\ref{table:concentration}.
\end{theorem}
\begin{proof}[Proof sketch (full proof in Appendix \ref{proof:thm3})]
These results follow by applying McDiarmid's inequality.
To apply it we need to show that RAM viewed as a function of $\XN$ has bounded differences.
We show that when replacing $\smash{X_i\in\XN}$ with $\smash{X_i'}$ the value of $\smash{D_f( \hat{Q}_Z^N \| P_Z)}$ changes by at most $\smash{O(N^{-1/2}\psi(N))}$.
Proof of this proceeds similarly to the one of Theorem \ref{thm:convergence-rate-general}.
\end{proof}

In practice it may not be possible to evaluate $\smash{D_f( \hat{Q}_Z^N \| P_Z)}$ analytically. 
We propose to use Monte-Carlo (MC) estimation since both densities $\hat{q}_N(z)$ and $p(z)$ are assumed to be known.
We consider importance sampling with proposal distribution ${\pi(z|\XN)}$, highlighting the fact that $\pi$ can depend on the sample $\XN$.
If $\pi(z|\XN) = p(z)$ this reduces to normal MC sampling. 
We arrive at the {\bf RAM-MC estimator} based on $M$ i.i.d.\:samples $\ZM:=\{Z_1,\dots,Z_M\}$ from $\pi(z|\XN)$:
\begin{align}
\label{eq:our-mc-estimate}
    \hat{D}^M_f( \hat{Q}_Z^N \| P_Z) :=
    \frac{1}{M}\sum_{m=1}^M f\left( \frac{\hat{q}_N(Z_m)}{p(Z_m)} \right) \frac{p(Z_m)}{\pi\left(Z_m|\XN\right)}.
\end{align}

\begin{theorem}[RAM-MC is unbiased and consistent]\label{thm:mc-variance}
$\E\bigl[\hat{D}^M_f( \hat{Q}_Z^N \| P_Z)\bigr]
=\E\bigl[D_f( \hat{Q}^N_{Z} \| P_Z )\bigr]$ for any proposal distribution $\pi$.
If $\pi(z|\XN) = p(z)$ or $\pi(z | \XN) = \hat{q}_N(z)$ then under mild assumptions$^\star$ on the moments of $q(Z|X)/p(Z)$
and denoting by ${\psi(N)}$ the rate given in Table~\ref{table:concentration}, we have
\begin{align*}
    \text{Var}_{\XN, \ZM} \bigl[\hat{D}^M_f( \hat{Q}_Z^N \| P_Z)\bigr] = 
    O\left(M^{-1}\right) + O\left( \psi(N)^2 \right).
\end{align*}
\end{theorem}
\begin{proof}[Proof sketch ($^\star$full statement and proof in Appendix \ref{appendix:full-statment-proof-mc})]
By the law of total variance, 
\begin{align*}
    \text{Var}_{\XN, \ZM} \bigl[\hat{D}^M_f\bigr] = 
    \mathbb{E}_{\XN} \bigl[\text{Var}\bigl[\hat{D}^M_f\, | \XN\bigr]\bigr] + \text{Var}_{\XN} \bigl[D_f( \hat{Q}_Z^N \| P_Z )\bigr].
\end{align*}
The first of these terms is ${O( M^{-1})}$ by standard results on MC integration, subject to the assumptions on the moments.
Using the fact that ${\text{Var}[Y] = \int_0^\infty\mathbb{P} ( |Y - \mathbb{E} Y| > \sqrt{t}) dt}$ for any random variable $Y$
we bound the second term by integrating the exponential tail bound of Theorem~\ref{thm:concentration}.
\end{proof}

Through use of the Efron-Stein inequality---rather than integrating the tail bound provided by McDiarmid's inequality---it is possible for some choices of $f$ to weaken the assumptions under which the $O(\psi(N)^2)$ variance is achieved: from uniform boundedness of $\smash{\chi^2(Q_{Z|X}\|P_Z)}$ to boundedness in expectation.
In general, a variance better than ${O(M^{-1})}$ is not possible using importance sampling. However, the constant and hence practical performance may vary significantly depending on the choice of $\pi$.
We note in passing that through Chebyshev's inequality, it is possible to derive confidence bounds for RAM-MC of the form similar to Theorem~\ref{thm:concentration}, but with an additional dependence on $M$ and worse dependence on $\delta$. 
For brevity we omit this.

\renewcommand{\arraystretch}{1}
\begin{table}
 \caption{Rate of bias for other estimators of $D_f(P,Q)$.}
 \label{table:convergence-other}
 \centering
 \begin{tabular}{c c c c c c c c c } 
 \toprule
 \multirow{2}{*}{$f$-divergence} & \multirow{2}{*}{KL} & \multirow{2}{*}{TV} & \multirow{2}{*}{$\chi^2$} & \multirow{2}{*}{$\text{H}^2$} & \multirow{2}{*}{JS} & \multicolumn{2}{c}{\thead{$D_{f_\beta}$}}  & \thead{$D_{f_\alpha}$} \\ [-0.8ex]
 & & & & & & $\scriptstyle{\frac{1}{2}<\beta<1}$ & $\scriptstyle{1<\beta<\infty}$ &
$\scriptstyle{-1<\alpha<1}$ \\
 \midrule
 \thead{Krishnamurthy et al. [22]} & - & - & - & - & - & - & - & $\scriptstyle{N^{-\frac{1}{2}} + N^{\frac{-3s}{2s + d}}}$ \\ 
 \thead{Nguyen et al. [28]} & $\scriptstyle{N^{-\frac{1}{2}}}$ & - & - & - & - & - & - & - \\ 
 \thead{Moon and Hero [26]} & $\scriptstyle{N^{-\frac{1}{2}}}$ & - & $\scriptstyle{N^{-\frac{1}{2}}}$ & $\scriptstyle{N^{-\frac{1}{2}}}$ & $\scriptstyle{N^{-\frac{1}{2}}}$ & $\scriptstyle{N^{-\frac{1}{2}}}$ & $\scriptstyle{N^{-\frac{1}{2}}}$ & $\scriptstyle{N^{-\frac{1}{2}}}$ \\ 
 \bottomrule
\end{tabular}
\end{table}

\subsection{Discussion: assumptions and summary}\label{subsection:discussion-assumptions}
All the rates in this section are independent of the dimension of the space $\mathcal{Z}$ over which the distributions are defined.
However the constants may exhibit some dependence on the dimension.
Accordingly, for fixed $N$, the bias and variance may generally grow with the dimension.

Although the data distribution $Q_X$ will generally be unknown, in some practical scenarios such as deep autoencoder models, $P_Z$ may be chosen by design and $Q_{Z|X}$ learned subject to architectural constraints.
In such cases, the assumptions of Theorems \ref{thm:convergence-rate-general} and \ref{thm:concentration} can be satisfied by making suitable restrictions (we conjecture also for Theorem~\ref{thm:fast-KL-rate}).
For example, suppose that ${P_Z}$ is  ${\mathcal{N}\left(0, I_d\right)}$ and ${Q_{Z|X}}$ is  ${\mathcal{N}\left( \mu(X), \Sigma(X)\right)}$ with $\Sigma$ diagonal. 
Then the assumptions hold if there exist constants $K, \epsilon > 0$ such that ${\| \mu(X)\| < K}$ and ${\Sigma_{ii}(X) \in [\epsilon, 1]}$ for all $i$ (see Appendix \ref{appendix:discussion-constraints}).
In practice, numerical stability often requires the diagonal entries of $\Sigma$ to be lower bounded by a small number (e.g. $10^{-6}$).
If $\mathcal{X}$ is compact (as for images) then such a $K$ is guaranteed to exist; if not, choosing $K$ very large yields an insignificant constraint.

Table~\ref{table:convergence-other} summarizes the rates of bias for some existing methods.
In contrast to our proposal, the assumptions of these estimators may in practice be difficult to verify.
For the estimator of \cite{krishnamurthy14icml}, both densities $p$ and $q$ must belong to the H\"older class of smoothness $s$, be supported on $[0,1]^d$ and satisfy $0<\eta_1 < p, q < \eta_2<\infty$ on the support for known constants $\eta_1, \eta_2$.
For that of \cite{nguyen10ratio}, the density ratio $p/q$ must satisfy $0<\eta_1 < p/q < \eta_2<\infty$ and belong to a function class $G$ whose \emph{bracketing entropy} (a measure of the complexity of a function class) is properly bounded. The condition on the bracketing entropy is quite strong and ensures that the density ratio is well behaved.
For the estimator of \cite{moon14ensemble}, both $p$ and $q$ must have the same bounded support and satisfy $0<\eta_1 < p, q < \eta_2<\infty$ on the support. $p$ and $q$ must have \emph{continuous bounded} derivatives of order $d$ (which is stronger than assumptions of [22]), and $f$ must have derivatives of order at least $d$.

In summary, the RAM estimator $D_f(\hat{Q}_Z^N \| P_Z)$ for $D_f(Q_Z \| P_Z)$ is \textbf{consistent} since it concentrates to its expectation $\E_{\XN}\bigl[D_f(\hat{Q}_Z^N \| P_Z)\bigr]$, which in turn converges to $D_f(Q_Z \| P_Z)$.
It is also \textbf{practical} because it can be efficiently estimated with Monte-Carlo sampling via RAM-MC.
\section{Empirical evaluation}\label{sec:experiments}

In the previous section we showed that our proposed estimator has a number of desirable theoretical properties.
Next we demonstrate its practical performance.
First, we present a synthetic experiment investigating the behaviour of RAM-MC in controlled settings where all distributions and divergences are known.
Second, we investigate the use of RAM-MC in a more realistic setting to estimate a divergence between the aggregate posterior $Q_Z$ and prior $P_Z$ in pretrained autoencoder models. 
For experimental details not included in the main text,
see Appendix \ref{appendix:empirical-evaluation-details}\footnote{
A python notebook to reproduce all experiments is available at \url{https://github.com/google-research/google-research/tree/master/f_divergence_estimation_ram_mc}.}.

\subsection{Synthetic experiments}\label{section:synth-exps}
\textbf{The data model.}
Our goal in this subsection is to test the behaviour of the RAM-MC estimator for various $d=\dim(\mathcal{Z})$ and $f$-divergences.
We choose a setting in which $Q^{\lambda}_Z$ parametrized by a scalar $\lambda$ and $P_Z$ are both $d$-variate normal distributions for $d\in\{1, 4, 16\}$.
We use RAM-MC to estimate $D_f(Q^\lambda_Z, P_Z)$, which can be computed analytically for the KL, $\chi^2$, and squared Hellinger divergences in this setting (see Appendix \ref{appendix:toy-exps}).
Namely, we take ${P_Z}$ and ${Q_X}$ to be standard normal distributions over $\Z=\R^d$ and $\X=\R^{20}$ respectively,
and $\smash{Z\sim Q^\lambda_{Z|X}}$ be a linear transform of $X$ plus a fixed isotropic Gaussian noise, with the linear function parameterized by $\lambda$.
By varying $\lambda$ we can interpolate between different values for $D_f(Q_Z^\lambda \| P_Z)$.

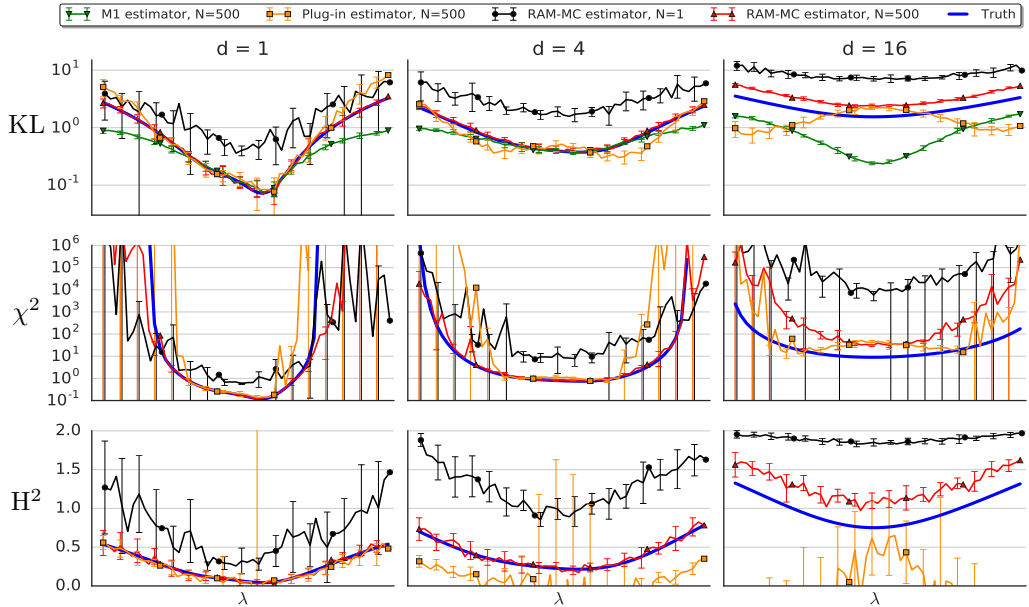
\begin{figure}
\begin{center}
\begin{tikzpicture}
\node[anchor=south west,inner sep=0] at (0,0) {\includegraphics[width=0.97\textwidth, height=0.615\textwidth]{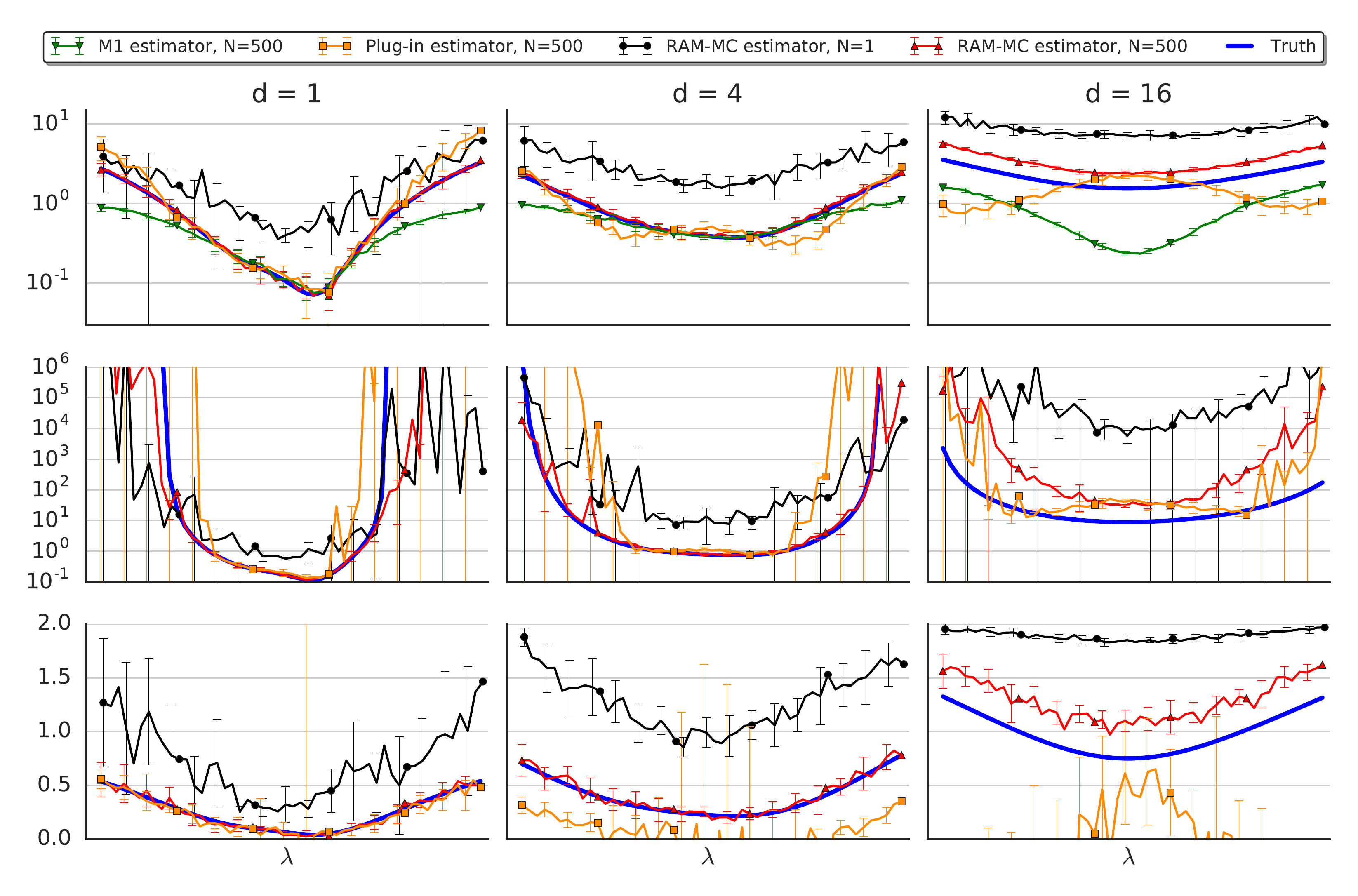}};
\node[rotate=0] at (0, 1.7) {$\mathrm{H}^2$};
\node[rotate=0] at (0, 4.2) {$\chi^2$};
\node[rotate=0] at (0, 6.7) {$\mathrm{KL}$};
\end{tikzpicture}
\end{center}
\caption{\label{fig:synthetic-exps}
(Section~\ref{section:synth-exps})
Estimating $D_f\bigl(\mathcal{N}(\mu_\lambda, \Sigma_\lambda),\, \mathcal{N}(0, I_d)\bigr)$ for various $f$, $d$, and parameters $\mu_\lambda$ and $\Sigma_\lambda$ indexed by $\lambda\in \R$.
Horizontal axis correspond to $\lambda\in[-2, 2]$,
columns to $d\in\{1, 4, 16\}$ and
rows to KL, $\chi^2$, and $\mathrm{H}^2$ divergences respectively.
{\bf \textcolor{blue}{Blue}} are true divergences, 
{\bf black} and {\bf \textcolor{red}{red}} are RAM-MC estimators~\eqref{eq:our-mc-estimate} for $N\in\{1, 500\}$ respectively,
{\bf \textcolor{darkgreen}{green}} are M1 estimator of~\citep{nguyen10ratio} and {\bf \textcolor{orange}{orange}} are plug-in estimates based on Gaussian kernel density estimation \citep{moon14ensemble}.
$N=500$ and $M=128$ in all the plots if not specified otherwise.
Error bars depict one standard deviation over 10 experiments.
}
\end{figure}

\textbf{The estimators.}
In Figure \ref{fig:synthetic-exps} we show the behaviour of RAM-MC with $N\,{\in}\,\{1, 500\}$ and $M{=}128$ compared to the ground truth as $\lambda$ is varied. 
The columns of Figure~\ref{fig:synthetic-exps} correspond to different dimensions $d\,{\in}\,\{1, 4, 16\}$, and rows to the $\KL$, $\chi^2$ and $\mathrm{H}^2$ divergences, respectively. 
We also include two baseline methods.
First, a plug-in method based on kernel density estimation \cite{moon14ensemble}.
Second, and only for the KL case, the M1 method of~\cite{nguyen10ratio} based on density ratio estimation.

\textbf{The experiment.}
To produce each plot, the following was performed 10 times, with the mean result giving the bold lines and standard deviation giving the error bars.
First, $N$ points $\XN$ were drawn from $Q_X$. 
Then $M{=}128$ points $\ZM$ were drawn from $\hat{Q}_Z^N$ and RAM-MC \eqref{eq:our-mc-estimate} was evaluated. 
For the plug-in estimator, the densities $\hat{q}(z)$ and $\hat{p}(z)$ were estimated by kernel density estimation with 500 samples from $Q_Z$ and $P_Z$ respectively using the default settings of the Python library {\texttt{scipy.stats.gaussian\_kde}}.
The divergence was then estimated via MC-sampling using $128$ samples from $Q_Z$ and the surrogate densities.
The M1~estimator involves solving a convex linear program in $N$ variables to maximize a lower bound on the true divergence, see \cite{nguyen10ratio} for more details.
Although the M1~estimator can in principle be used for arbitrary $f$-divergences, its implementation requires hand-crafted derivations that are supplied only for the $\KL$ in \cite{nguyen10ratio}, which are the ones we use.

\textbf{Discussion.}
The results of this experiment empirically support Proposition \ref{prop:upper-bound} and Theorems \ref{thm:fast-KL-rate}, \ref{thm:convergence-rate-general}, and~\ref{thm:mc-variance}:
(i) in expectation, RAM-MC upper bounds the true divergence; (ii) by increasing $N$ from 1 to 500 we clearly decrease both the bias and the variance of RAM-MC.
When the dimension $d$ increases, the bias for fixed $N$ also increases.
This is consistent with the theory in that, although the rates are independent of $d$, the constants are not.
We note that by side-stepping the issue of density estimation, RAM-MC performs favourably compared to the plug-in and M1 estimators, more so in higher dimensions ($d=16$).
In particular, the shape of the RAM-MC curve follows that of the truth for each divergence, while that of the plug-in estimator does not for larger dimensions.
In some cases the plug-in estimator can even take negative values because of the large variance.

\subsection{Real-data experiments}
\label{sec:exp_wae}
\textbf{The data model.}
To investigate the behaviour of RAM-MC in a more realistic setting, we consider Variational Autoencoders (VAEs) and Wasserstein Autoencoders (WAEs) \cite{kingma2013auto, tolstikhin2017wasserstein}.
Both models involve learning an \emph{encoder} $\smash{Q^\theta_{Z|X}}$ with parameter $\theta$ mapping from high dimensional data to a lower dimensional latent space and decoder mapping in the reverse direction.
A prior distribution ${P_Z}$ is specified, and the 
optimization objectives of both models are of the form ``reconstruction + distribution matching penalty''.
The penalty of the VAE was shown by \cite{hoffman2016elbo} to be equivalent to $\smash{\KL(Q^\theta_Z \| P_Z) + I(X,Z)}$ where $I(X,Z)$ is the mutual information of a sample and its encoding.
The WAE penalty is ${D(Q^\theta_Z \| P_Z)}$ for any divergence $D$ that can practically be estimated.
Following \cite{tolstikhin2017wasserstein}, we trained models using the Maximum Mean Discrepency (MMD), a kernel-based distance on distributions, and a divergence estimated using a GAN-style classifier leading to WAE-MMD and WAE-GAN respectively \cite{gretton2012kernel, goodfellow2014generative}.
For more information about VAE and WAE, see Appendix \ref{appendix:intro-vae-wae}.

\begin{figure}
\begin{center}
\includegraphics[width=1.\textwidth, height=0.615\textwidth]{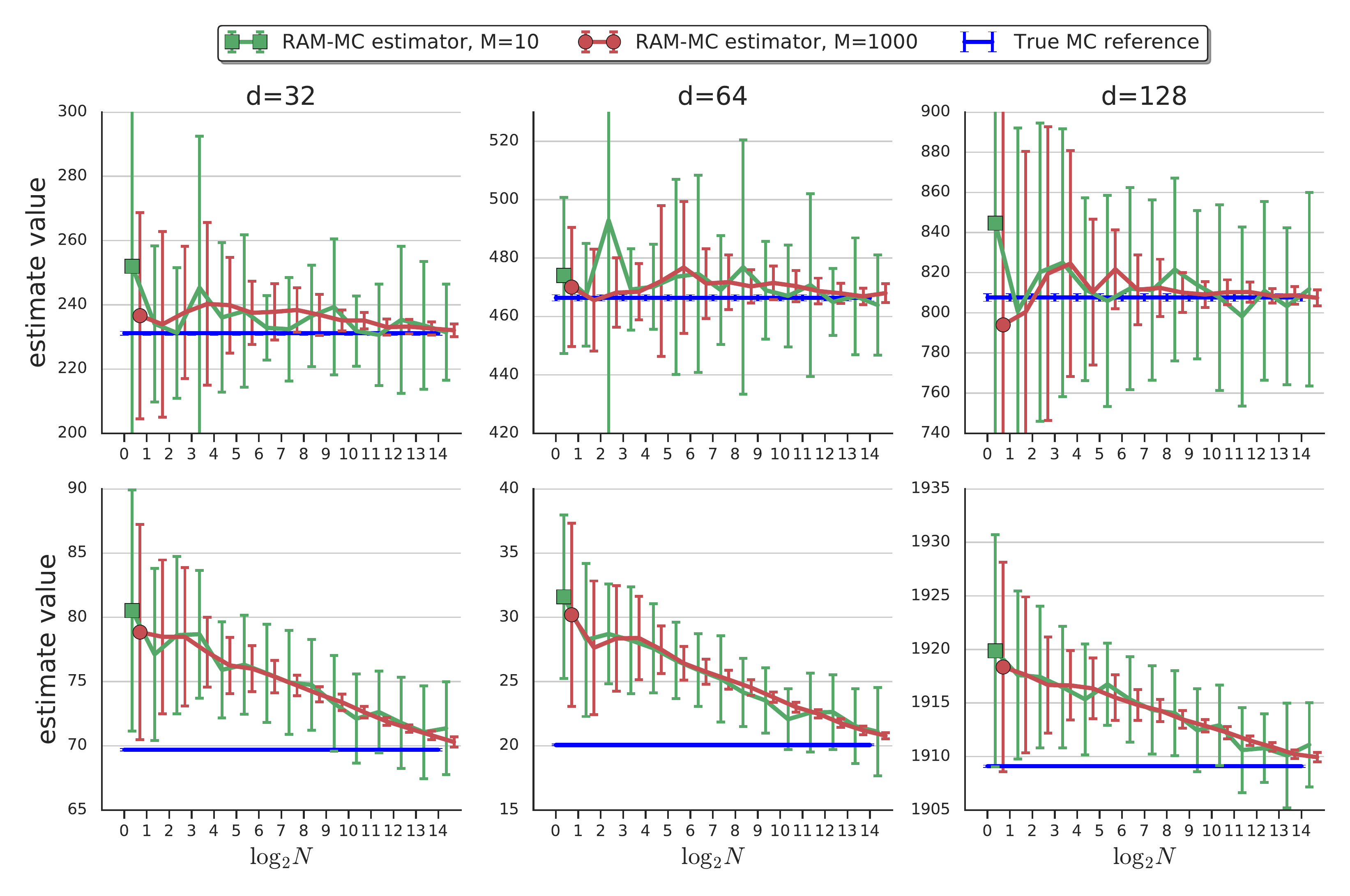}
\end{center}
\caption{\label{fig:real-exps}
(Section~\ref{sec:exp_wae}) Estimates of $KL(Q_Z^\theta \| P_Z)$ for pretrained autoencoder models with RAM-MC as a function of $N$ for $M{=}10$ ({\bf \textcolor{green!65!blue}{green}}) and $M{=}1000$ ({\bf \textcolor{red}{red}}) compared to an accurate MC estimate of the ground truth ({\bf\textcolor{blue}{blue}}).
Lines and error bars represent means and standard deviations over 50 trials.
}
\end{figure}

\textbf{The experiment.}
We consider models pre-trained on the \emph{CelebA} dataset \cite{liu2015faceattributes}, 
and use them to evaluate the RAM-MC estimator as follows.
We take the test dataset as the ground-truth $Q_X$, and embed it into the latent space via the trained encoder.
As a result, we obtain a ${\sim}{20}\text{k}$-component Gaussian mixture for $Q_Z$, the \emph{empirical aggregate posterior}. 
Since $Q_Z$ is a finite---not continuous--- mixture, the true $D_f(Q_Z\|P_Z)$ can be estimated using a large number of MC samples (we used $10^4$).
Note that this is very costly and involves evaluating $2\cdot 10^4$ Gaussian densities for each of the $10^4$ MC points.
We repeated this evaluation 10 times and report means and standard deviations.
RAM-MC is evaluated using $N \in \{2^0, 2^1,\ldots, 2^{14}\}$ and $M \in \{10, 10^3\}$.
For each combination $(N,M)$, RAM-MC was computed 50 times with the means plotted as bold lines and standard deviations as error bars.
In Figure~\ref{fig:real-exps} we show the result of performing this for the KL divergence on six different models.
For each dimension $d\in\{32, 64, 128\}$, we chose two models from the classes (VAE, WAE-MMD, WAE-GAN). 
See Appendix~\ref{appendix:real-data-experiments-additional} for further details and similar plots for the $H^2$-divergence.

\textbf{Discussion.}
The results are encouraging. 
In all cases RAM-MC achieves a reasonable accuracy with $N$ relatively small, even for the bottom right model where the true KL divergence ($\approx 1910$) is very big.
We see evidence supporting Theorem~\ref{thm:mc-variance}, which says that the variance of RAM-MC is mostly determined by the smaller of $\psi(N)$ and $M$:
when $N$ is small, the variance of RAM-MC does not change significantly with $M$, 
however when $N$ is large, increasing $M$ significantly reduces the variance. 
Also we found there to be two general modes of behaviour of RAM-MC across the six trained models we considered. 
In the bottom row of Figure~\ref{fig:real-exps} we see that the decrease in bias with $N$ is very obvious, supporting Proposition~\ref{prop:upper-bound} and Theorems \ref{thm:fast-KL-rate} and \ref{thm:convergence-rate-general}.
In contrast, in the top row it is less obvious, because the comparatively larger variance for $M{=}10$ dominates reductions in the bias.
Even in this case, both the bias and variance of RAM-MC with $M{=}1000$ become negligible for large $N$.
Importantly, the behaviour of RAM-MC does not degrade in higher dimensions.

The baseline estimators (plug-in \cite{moon14ensemble} and M1~\cite{nguyen10ratio}) perform so poorly that we decided not to include them in the plots (doing so would distort the $y$-axis scale).
In contrast, even with a relatively modest $N{=}2^8$ and $M{=}1000$ samples, RAM-MC behaves reasonably well in all cases.

\section{Applications: total correlation, entropy, and mutual information estimates}\label{sec:applications}
In this section we describe in detail some direct consequences of our new estimator and its guarantees.
Our theory may also apply to a number of machine learning domains where estimating entropy, total correlation or mutual information is either the final goal or part of a broader optimization loop.
\paragraph{Total correlation and entropy estimation.}
The differential entropy, which is defined as $H(Q_Z)= -\int_{\mathcal{Z}} q(z) \log q(z)  dz$, is often a quantity of interest in machine learning.
While this is intractable in general, straightforward computation shows that for \emph{any} $P_Z$
{\addtolength{\abovedisplayskip}{-0.5mm}
\addtolength{\belowdisplayskip}{-0.5mm}
\begin{align*}
    H(Q_Z) - \mathbb{E}_{\XN} H(\hat{Q}_Z^N) = \mathbb{E}_{\XN} \KL[\hat{Q}_Z^N \| P_Z] -  \KL[Q_Z \| P_Z].
\end{align*}}%
Therefore, our results provide sufficient conditions under which $\smash{H(\hat{Q}_Z^N)}$ converges to $\smash{H(Q_{Z})}$ and concentrates to its mean.
We now examine some consequences for Variational Autoencoders (VAEs).

Total Correlation is considered by \cite{chen2018isolating},
$
TC(Q_Z) := \KL[Q_Z \| \prod_{i=1}^{d_Z} Q_{Z_i}] =     \sum_{i=1}^{d_Z}H(Q_{Z_i}) - H(Q_Z)
$
where $Q_{Z_i}$ is the $i$th marginal of $Q_Z$.
This is added to the VAE loss function to encourage $Q_Z$ to be factorized, resulting in the $\beta$-TC-VAE algorithm.
By the second equality above, estimation of TC can be reduced to estimation of $H(Q_Z)$ (only slight modifications are needed to treat $H(Q_{Z_i})$).

Two methods are proposed in \cite{chen2018isolating} for estimating $\smash{H(Q_Z)}$, both of which assume a finite dataset of size $D$.
One of these, named \emph{Minibatch Weighted Sample} (MWS), coincides with $\smash{H(\hat{Q}_Z^N) + \log D}$ estimated with a particular form of MC sampling.
Our results therefore imply \emph{inconsistency} of the MWS method due to the constant $\log D$ offset. 
In the context of \cite{chen2018isolating} this is not actually problematic since a constant offset does not affect gradient-based optimization techniques.
Interestingly, although the derivations of \cite{chen2018isolating} suppose a data distribution of finite support, our results show that minor modifications result in an estimator suitable for both finite and infinite support data distributions.

\paragraph{Mutual information estimation.}
The mutual information (MI) between variables with joint distribution $\smash{Q_{Z,X}}$ is defined as $\smash{I(Z, X) := \KL\left[Q_{Z,X} \| Q_Z Q_X \right] = \E_{X} \KL\left[Q_{Z|X} \| Q_Z \right]}$.
Several recent papers have estimated or optimized this quantity in the context of autoencoder architectures, coinciding with our setting \cite{dieng2018avoiding, hoffman2016elbo, alemi2017fixing, oord2018representation}. In particular, \cite{poolevariational} propose the following estimator based on replacing $Q_Z$ with $\smash{\hat{Q}_{Z}^N}$, proving it to be a lower bound on the true MI:
{\addtolength{\abovedisplayskip}{-0.6mm}
\addtolength{\belowdisplayskip}{-0.6mm}
\begin{align*}\textstyle
    I_{TCPC}^N(Z,X) = \mathbb{E}_{\XN}\Big[\frac{1}{N} \sum_{i=1}^N \KL[ Q_{Z|X_i} \| \hat{Q}_{Z}^N ]\Big] \leq I(Z,X).
\end{align*}}%
The gap can be written as
$\smash{I(Z,X) - I_{TCPC}^N(Z,X) = \E_{\XN} \KL[ \hat{Q}_{Z}^N \| P_{Z} ] - \KL[ Q_{Z} \| P_Z ]}$
where $P_Z$ is \emph{any} distribution. 
Therefore, our results also provide sufficient conditions under which $\smash{I^N_{TCPC}}$ converges and concentrates to the true mutual information.
\section{Conclusion}\label{sec:conclusion}
We introduced a practical estimator for the $\smash{f}$-divergence $D_f(Q_Z\|P_Z)$ where $Q_Z = \int Q_{Z|X}dQ_X$, samples from $Q_X$ are available, and $P_Z$ and $Q_{Z|X}$ have known density.
The RAM estimator is based on approximating the true $Q_Z$ with data samples as a random mixture via $\smash{\hat{Q}^N_{Z}=\frac{1}{N}\sum_{n} Q_{Z|X_n}}$.
We denote by RAM-MC the estimator version where $\smash{D_f(\hat{Q}^N_Z\|P_Z)}$ is estimated with MC sampling.
We proved rates of convergence and concentration for both RAM and RAM-MC, in terms of sample size $N$ and MC samples $M$ under a variety of choices of $\smash{f}$.
Synthetic and real-data experiments strongly support the validity of our proposal in practice, and our theoretical results provide guarantees for methods previously proposed heuristically in existing literature.

Future work will investigate the use of our proposals for optimization loops, in contrast to pure estimation.
When $\smash{Q^\theta_{Z|X}}$ depends on parameter $\theta$ and the goal is to minimize $\smash{D_f(Q_Z^\theta \| P_Z)}$ with respect to $\theta$, RAM-MC provides a practical surrogate loss that can be minimized using stochastic gradient methods.

\subsection*{Acknowledgements}
Thanks to Alessandro Ialongo, Niki Kilbertus, Luigi Gresele, Giambattista Parascandolo, Mateo Rojas-Carulla and the rest of Empirical Inference group at the MPI, and Ben Poole, Sylvain Gelly, Alexander Kolesnikov and the rest of the Brain Team in Zurich for stimulating discussions, support and advice.

\small

\bibliographystyle{plain}
\bibliography{main}

\appendix
\newpage

\section{$f$ for divergences considered in this paper}\label{appendix:f-fns}

One of the useful properties of $f$-divergences that we make use of in the proofs of Theorems~\ref{thm:convergence-rate-general} and \ref{thm:concentration} is that for any constant $c$, replacing $f(x)$ by $f(x) + c(x-1)$ does not change the divergence $D_f$. 
It is often convenient to work with $f_0(x) := f(x) - f'(1)(x-1)$ which is decreasing on $(0, 1)$ and increasing on $(1, \infty)$ and satisfies $f'_0(1)=0$.

In Table \ref{table:f-fns} we list the forms of the function $f_0$ for each of the divergences considered in this paper.

{
\renewcommand{\arraystretch}{2}
\begin{table}
 \caption{$f$ corresponding to divergences referenced in this paper.}
 \label{table:f-fns}
 \centering
 \begin{tabular}{c c} 
 \toprule
 $f$-divergence & $f_0(x)$ \\
 \midrule
 KL & $x \log x - x + 1$\\
 TV & $\frac{1}{2}|1-x|$\\
 $\chi^2$ & $x^2 - 2x$\\
 $\text{H}^2$ & $2(1-\sqrt{x})$\\
 JS & $(1+x)\log(\frac{2}{1+x}) + x\log x$\\
 $D_{f_\beta}$, $\beta > 0,$ $\beta\not=\frac{1}{2}$ & $\frac{1}{1-\frac{1}{\beta}}\left[ (1+x^\beta)^{\frac{1}{\beta}} - 2^{\frac{1}{\beta}-1}(1+x) \right]$\\
 $D_{f_\alpha}$, $-1<\alpha < 1$ & $\frac{4}{1-\alpha^2}\left( 1 - x^{\frac{1+\alpha}{2}} \right) - \frac{2(x-1)}{\alpha-1}$ \\
 \bottomrule
\end{tabular}
\end{table}
}
\section{Proofs}\label{appendix:proofs}

\subsection{Proof of Proposition \ref{prop:upper-bound}}\label{proof:prop1}

\setcounter{proposition}{0}
\begin{proposition}
Let $M \leq N$ be integers. Then
\begin{align*}
    D_f(Q_Z \| P_Z) \ \leq 
    \mathbb{E}_{\mathbf{X}^N \sim Q_{X}^N} D_f( \hat{Q}_Z^N \| P_Z) \  \leq \ \mathbb{E}_{\mathbf{X}^M \sim Q_{X}^M} D_f( \hat{Q}_Z^M \| P_Z).
\end{align*}
\end{proposition}
\begin{proof}
Observe that $\mathbb{E}_{\mathbf{X}^N} \hat{Q}_Z^N = Q_Z$. Thus,
\begin{align*}
    D_f(Q_Z \| P_Z) &= \int f\left(\frac{\E_{\XN}\hat{q}_N(z)}{p(z)}\right) dP_Z(z) \\
    &\leq \E_{\XN} \int f\left(\frac{\hat{q}_N(z)}{p(z)}\right) dP_Z(z)\\
    &=\mathbb{E}_{\mathbf{X}^N \sim P_{X}^N} D_f( \hat{Q}_Z^N \| P_Z),
\end{align*}
where the inequality follows from convexity of $f$.

To see that $\mathbb{E}_{\mathbf{X}^N \sim P_{X}^N} D_f( \hat{Q}_Z^N \| P_Z) \leq \mathbb{E}_{\mathbf{X}^M \sim P_{X}^N} D_f( \hat{Q}_Z^M \| P_Z)$ for $N \geq M$,
let $I \subseteq \{1, \ldots, N\}$, $|I| = M$ and write
\begin{align*}
    \hat{Q}_Z^I = \frac{1}{M} \sum_{i \in I} Q_{Z | X_i}.
\end{align*}

Letting $I$ be a random subset chosen uniformly \emph{without replacement},
observe that for any fixed $I$, $\mathbf{X}^I \sim \mathbb{P}_X^M$ (with the randomness coming from $\mathbf{X}^N \sim \mathbb{P}_X^N$). Thus

\begin{align*}
    \hat{Q}_Z^N &= \frac{1}{N} \sum_{i=1}^N Q_{Z|X_i} \\
    &= \mathbb{E}_I \frac{1}{M} \sum_{i \in I} Q_{Z|X_i} \\
    &= \mathbb{E}_I \hat{Q}_Z^I
\end{align*}

and so again by convexity of $f$ we have that

\begin{align}
    \mathbb{E}_{\mathbf{X}^N \sim P_{X}^N} D_f( \hat{Q}_Z^N \| P_Z) &\leq  \mathbb{E}_{\mathbf{X}^N} \mathbb{E}_I D_f( \hat{Q}_Z^I \| P_Z) \\
    &= \mathbb{E}_{\mathbf{X}^M} D_f(\hat{Q}_Z^M \| P_Z)
\end{align}
with the last line following from the observation that $\mathbf{X}^I \sim \mathbb{P}_X^M$.
\end{proof}

\subsection{Proof of Theorem \ref{thm:fast-KL-rate}}\label{appendix:subsec:thm1}

\begin{lemma}\label{lemma:hilbertian-triangle}
Suppose that $D_f^{\frac{1}{2}}$ satisfies the triangle inequality.
Then for any $\lambda>0$,
\begin{align*}
    D_f\left(\hat{Q}^N_Z \| P_Z\right) - D_f\left(Q_{Z} \| P_Z\right) \leq (1+\lambda) D_f\left(\hat{Q}^N_Z \| Q_Z \right) +  \frac{1}{\lambda} D_f\left(Q_{Z} \| P_Z \right)
\end{align*}
If, furthermore, $\E_{\XN}\left[ D_f\left(\hat{Q}^N_Z \| Q_Z\right)\right] = O\left( \frac{1}{N^k} \right)$ 
for some $k>0$, 
then
\begin{align*}
    \E_{\XN}\left[ D_f\left(\hat{Q}^N_Z \| P_Z\right) \right] - D_f\left(Q_{Z} \| P_Z\right) = O\left( \frac{1}{N^{k/2}} \right)
\end{align*}
\end{lemma}
\begin{proof}
The first inequality follows from the triangle inequality for $D_f^{\frac{1}{2}}$ on $\hat{Q}^N_Z$ and $P_Z$, and the fact that $2\sqrt{ab} \leq \lambda a + \frac{b}{\lambda}$ for ${a, b, \lambda>0}$.
The second inequality follows from the first by taking $\lambda = N^{-\frac{k}{2}}$.
\end{proof}

\setcounter{theorem}{0}
\begin{theorem}[Rates of the bias]
If
$\E_{X\sim Q_X}\bigl[\chi^2\bigl(Q_{Z|X}, Q_Z\bigr)\bigr]$ and
$\KL\left( Q_{Z} \| P_Z\right)$ are finite then the bias ${\E_{\XN}\bigl[D_f( \hat{Q}_Z^N \| P_Z)\bigr] - D_f\left( Q_{Z} \| P_Z\right)}$ decays with rate as given in the first row of Table~\ref{table:convergence}.
\end{theorem}

\begin{proof}
To begin, observe that 
\begin{align*}
    \E_{\XN}\left[\chi^2\bigl(\hat{Q}^N_Z, Q_Z\bigr)\right]
    &= \E_{\XN}\E_{Q_Z}\left[\left(\frac{\hat{q}_N(z)}{q(z)} - 1\right)^2 \right]\\
    &=\E_{Q_Z} \V_{\XN} \left[ \frac{1}{N} \sum_{n=1}^N\frac{q(z|X_n)}{q(z)}\right] \\
    &= \frac{1}{N} \E_{Q_Z}\V_X \left[ \frac{q(z|X)}{q(z)} \right]\\
    &= \frac{1}{N}\E_{X}\left[\chi^2\bigl(Q_{Z|X}, Q_Z\bigr) \right]
\end{align*}

where the introduction of the variance operator follows from the fact that $\E_{X_N}\left[ \frac{\hat{q}_N(z)}{q(z)} \right] = 1$.

For the $\KL$-divergence, using the fact that $\KL \leq \chi^2$ (Lemma 2.7 of \cite{tsybakov2009}) yields

\begin{align*}
    \E_{\XN}\left[\KL\left( \hat{Q}^N_Z \| P_Z\right)\right] - \KL\left( Q_{Z} \| P_Z\right) &=\E_{\XN}\left[\KL\left( \hat{Q}^N_Z \| Q_Z\right)\right]\\
    &\leq \E_{\XN}\left[\chi^2\bigl(\hat{Q}^N_Z, Q_Z\bigr) \right]\\
    &= \frac{1}{N}\E_{X}\left[\chi^2\bigl(Q_{Z|X}, Q_Z\bigr)\right]\\
    &=O\left(\frac{1}{N}\right),
\end{align*}
where the first equality can be verified by using the definition of $\KL$ and the fact that $Q_Z = \E_{\XN}\hat{Q}^N_Z$.

For Total Variation, we have
\begin{align*}
    \E_{\XN}\left[\TV\left( \hat{Q}^N_Z \| P_Z\right)\right] - \TV\left( Q_{Z} \| P_Z\right) &\leq\E_{\XN}\left[\TV\left( \hat{Q}^N_Z \| Q_Z\right)\right]\\
    &\leq \frac{1}{\sqrt{2}} \sqrt{\E_{\XN}\left[\KL\left( \hat{Q}^N_Z \| Q_Z\right) \right]}\\
    &=O\left(\frac{1}{\sqrt{N}}\right),
\end{align*}
where the first inequality holds since $\TV$ is a metric and thus obeys the triangle inequality, and the second inequality follows by Pinsker's inequality combined with concavity of $\sqrt{x}$ (Lemma 2.5 of \cite{tsybakov2009}).

For $D_{f_\beta}$ (including Jenson-Shannon) using the fact that $D_{f_\beta}^{1/2}$ satisfies the triangular inequality, we apply the second part of Lemma~\ref{lemma:hilbertian-triangle}
in combination with the fact that
$D_{f_\beta}\left(\hat{Q}^N_Z \| Q_Z\right) \leq \psi(\beta) \ \TV\left( \hat{Q}^N_Z \| Q_Z \right)$ for some scalar $\psi(\beta)$ (Theorem 2 of \cite{osterreicher2003new}) to obtain

\begin{align*}
    \E_{\XN}\left[D_{f_\beta}\left( \hat{Q}^N_Z \| P_Z\right)\right] - D_{f_\beta}\left( Q_{Z} \| P_Z\right) \leq O\left(\frac{1}{N^{1/4}}\right).
\end{align*}

Although the squared Hellinger divergence is a member of the $f_\beta$-divergence family, we can use the tighter bound $\Hsq\left( \hat{Q}^N_Z \| Q_Z\right) \leq KL\left( \hat{Q}^N_Z \| Q_Z\right)$ (Lemma 2.4 of \cite{tsybakov2009}) in combination with Lemma~\ref{lemma:hilbertian-triangle} to obtain
\begin{align*}
    \E_{\XN}\left[\Hsq\left( \hat{Q}^N_Z \| P_Z\right)\right] - \Hsq\left( Q_{Z} \| P_Z\right) \leq O\left(\frac{1}{\sqrt{N}}\right).
\end{align*}
\end{proof}

\subsection{Upper bounds of f}\label{appendix:subsubsec:f-upper-bounds}

We will make use of the following lemmas in the proof of Theorem \ref{thm:convergence-rate-general} and \ref{thm:concentration}.

\begin{lemma}\label{lemma:concave-upper-bound-kl} 
Let $f_0(x)=x\log x - x +1$, corresponding to $D_{f_0} = \KL$.
Write $g(x) = f_0'^2(x) = \log^2(x)$.

For any $0< \delta < 1$, the function
\begin{align*}
    h_{\delta}(x) := \begin{cases} 
    g(\delta) + x g'(e) & \: x \in [0, e]\\
    g(\delta) + e g'(e) + g(x) - g(e) & \: x \in [e, \infty)
    \end{cases}
\end{align*}
is an upper bound of $g(x)$ on $[\delta, \infty)$, and is concave and non-negative on $[0, \infty)$.
\end{lemma}

\begin{proof}

First observe that $h_{\delta}$ is concave.
It has continuous first and second derivatives:
\begin{align*}
    h_{\delta}'(x) = \begin{cases} 
    g'(e) & \: x \in [0, e]\\
    g'(x) & \: x \in [e, \infty)
    \end{cases}
    &&
    h_{\delta}''(x) = \begin{cases} 
    0 & \: x \in [0, e]\\
    g''(x) & \: x \in [e, \infty)
    \end{cases}
\end{align*}
Note that $g''(x) = \frac{2}{x^2} - \frac{2 \log(x)}{x^2} \leq 0$ for $x \geq e$ and $g''(e) = 0$.
Therefore $h''_{\delta}(x)$ has non-positive second derivative on $[0, \infty)$ and is thus concave on this set.

To see that $h_{\delta}(x)$ is an upper bound of $g(x)$ for $x \in [\delta, \infty)$, use the fact that $g'(x) = \frac{2\log(x)}{x}$ and observe that
\begin{align*}
    h_{\delta}(x) - g(x)
    &= \begin{cases} 
    \log^2(\delta) + \frac{2x}{e} - \log^2(x)& \: x \in [\delta, e]\\
    \log^2(\delta) + 1& \: x \in [e, \infty)
    \end{cases}
    \ > 0.
\end{align*}

To see that $h_{\delta}(x)$ is non-negative on $[0, \infty)$, note that $h_{\delta}(x) > g(x) \geq 0$ on $[\delta, \infty)$. 
Moreover, $g'(e) = 2/e > 0$, and so for $x \in [0, \delta]$ we have that $h_{\delta}(x) = g(\delta) + 2x/e \geq g(\delta) \geq 0$.
\end{proof}

\begin{lemma}\label{lemma:upper-bound-hellinger}
Let $f_0(x) = 2(1 -\sqrt{x})$ corresponding to the square of the Hellinger distance. 
Write $g(x) = f'^2_0(x) = (1-\frac{1}{\sqrt{x}})^2$.
For any $0<\delta<1$, the function
\begin{align*}
    h_\delta(x) = \frac{1}{\delta}(x-1)^2 
\end{align*}
is an upper bound of $g(x)$ on $[\delta, \infty)$.
\end{lemma}
\begin{proof}
For $x=1$, we have $g(1)=h_\delta(1)$. 
For $x\not=1$,
\begin{align*}
    0 &\leq \frac{1}{\delta}(x-1)^2 - (1-\frac{1}{\sqrt{x}})^2 \\
    \iff \sqrt{\delta} &\leq \frac{x-1}{1-\frac{1}{\sqrt{x}}}
\end{align*}
If $x \in [\delta, 1)$ then
\begin{align*}
    \frac{x-1}{1-\frac{1}{\sqrt{x}}} &= \sqrt{x} \cdot \frac{\frac{1}{\sqrt{x}} - \sqrt{x}}{\frac{1}{\sqrt{x}}-1} \geq \sqrt{x} \geq \sqrt{\delta}.
\end{align*}
If $x \in (1, \infty)$ then
\begin{align*}
    \frac{x-1}{1-\frac{1}{\sqrt{x}}} &= \sqrt{x} \cdot \frac{\sqrt{x} - \frac{1}{\sqrt{x}}}{1-\frac{1}{\sqrt{x}}} \geq \sqrt{x} \geq \sqrt{\delta}.
\end{align*}
Thus $g(x) \leq h_\delta(x)$ for $x\in[\delta, \infty)$.
\end{proof}

\begin{lemma}\label{lemma:upper-bound-alpha}
Let $f_0(x) = \frac{4}{1-\alpha^2}\left(1 -x^{\frac{1+\alpha}{2}}\right) - \frac{2(x-1)}{\alpha-1}$ corresponding to the $\alpha$-divergence with $\alpha \in (-1,1)$. 
Write $g(x) = f'^2_0(x) = \frac{4}{(\alpha -1)^2} \left(x^\frac{\alpha-1}{2} - 1\right)^2$.
For any $0<\delta<1$, the function
\begin{align*}
    h_\delta(x) = \frac{4\left(\delta^\frac{\alpha-1}{2} - 1\right)^2}{(\alpha - 1)^2 (\delta- 1)^2}\cdot (x-1)^2
\end{align*}
is an upper bound of $g(x)$ on $[\delta, \infty)$.
\end{lemma}
\begin{proof}
For $x=1$, we have $g(1)=h_\delta(1)$. 
Consider now the case that $x\geq\delta$ and $x\not=1$. 
Since $0<\delta <1$, we have that $1-\delta > 0$.
And because $(\alpha-1)/2 \in (-1,0)$, we have that $\delta^{\frac{\alpha - 1}{2}} - 1 > 0$.
It follows by taking square roots that
\begin{align*}
    &g(x) \leq h_\delta(x) \\
    \iff& d(x) := \frac{x^{\frac{\alpha -1}{2}} -1}{1-x} \leq \frac{\delta^{\frac{\alpha -1}{2}} -1}{1-\delta} 
\end{align*}
Now, $d(x)$ is non-increasing for $x>0$. Indeed,
\begin{align*}
    d'(x) = \frac{-1}{(1-x)^2} \left[ 1 - \frac{3 - \alpha}{2}x^{\frac{\alpha -1}{2}} + \frac{1-\alpha}{2} x^{\frac{\alpha-3}{2}} \right]
\end{align*}
and it can be shown by differentiating that the term inside the square brackets attains its minimum at $x=1$ and is therefore non-negative. Since $(1-x)^2 \geq 0$ it follows that $d'(x) \leq 0$ and so $d(x)$ is non-increasing.
From this fact it follows that $d(x)$ attains its maximum on $x \in [\delta, \infty)$ at $x=\delta$, and thus the desired inequality holds. 
\end{proof}

\begin{lemma}\label{lemma:upper-bound-JS}
Let $f_0(x) = \left(1+x\right) \log 2 + x\log x - \left( 1 + x\right) \log \left(1+x\right)$ corresponding to the Jensen-Shannon divergence.
Write $g(x) = f'^2_0(x) = \log^2 2 + \log^2\left( \frac{x}{1+x} \right) + 2\log 2 \log\left( \frac{x}{1+x} \right)$.
For $0< \delta < 1$, the function
\begin{align*}
    h_\delta(x) = g(\delta) + 4\log^2 2
\end{align*}
is an upper bound of $g(x)$ on $[\delta, \infty)$.
\end{lemma}
\begin{proof}
For $x\geq1$,  $\frac{x}{x+1} \in [0.5, 1)$ and so $\log\left( \frac{x}{1+x} \right) \in \left[-\log2, 0\right)$.
Therefore $g(x) \in \left(0, 4\log^2 2\right]$ for $x>1$.
It follows that for any value of $\delta$, $h_\delta(x) \geq g(x)$ for $x\geq1$.
$f'_0(1)=0$ and by differentiating again it can be shown that $f''_0(x) > 0$ for $x\in(0,1)$.
Thus $f'_0(x)<0$ and is increasing on $(0,1)$ and so $g(x) > 0$ and is decreasing on $(0,1)$.
Thus $h_\delta(x) > g(\delta) \geq g(x)$ for $x \in [\delta, 1)$.
\end{proof}

\begin{lemma}\label{lemma:upper-bound-f-beta}
Let $f_0(x) = \frac{1}{1-\frac{1}{\beta}}\left[ \left(1+x^\beta\right)^\frac{1}{\beta}  - 2^{\frac{1}{\beta} - 1}(1+x)\right]$
corresponding to the ${f_\beta}$-divergence introduced in \cite{osterreicher2003new}.
We assume $\beta \in \left( \frac{1}{2}, \infty\right) \setminus \{1\}$.
Write $g(x) = f'^2_0(x) = \left(\frac{\beta}{1-\beta}\right)^2\left[ \left(1+x^{-\beta}\right)^\frac{1-\beta}{\beta}  - 2^{\frac{1}{\beta} - 1}\right]^2$.

If $\beta \in \left(\frac{1}{2}, 1\right)$, then $\lim_{x\to \infty}g(x)$ exists and is finite and for any $0<\delta<1$, we have that $h_\delta(x) := g(\delta) + \lim_{x\to \infty}g(x) \geq g(x)$ for all $x\in[\delta, \infty)$.

If $\beta \in \left(1, \infty \right)$, then $\lim_{x\to0}g(x)$ and $\lim_{x\to \infty}g(x)$ both exist and are finite, and $g(x) \leq \max\{\lim_{x\to0}g(x), \lim_{x\to \infty}g(x)\}$ for all $x\in[0, \infty)$.
\end{lemma}
\begin{proof}
For any $\beta \in \left( \frac{1}{2}, \infty\right) \setminus \{1\}$, we have that $f''_0(x) = \frac{\beta}{(1-\beta)^2} \left[ 
\frac{1}{x^{\beta+1}} \left( 1 + x^{-\beta}\right)^{\frac{1-2\beta}{\beta}}\right] > 0$ for $x>0$.
Since $f'_0(1)=0$, it follows that $f'_0(x)$ is increasing everywhere, negative on $(0,1)$ and positive on $(1,\infty)$.
It follows that $g(x)$ is decreasing on $(0,1)$ and increasing on $(1,\infty)$.
$\beta > 0$ means that $1+x^{-\beta} \to 1$ as $x\to \infty$. Hence $g(x)$ is bounded above and increasing in $x$, thus $\lim_{x\to\infty} g(x)$ exists and is finite.

For $\beta \in (\frac{1}{2}, 1)$, $\frac{1-\beta}{\beta} > 0$. 
It follows that $\left(1+x^{-\beta}\right)^\frac{1-\beta}{\beta}$ grows unboundedly as $x \to 0$, and hence so does $g(x)$.
Since $g(x)$ is decreasing on $(0,1)$, for any $0<\delta<1$ we have that $h_\delta(x)\geq g(x)$ on $(0,1)$.
Since $g(x)$ is increasing on $(1, \infty)$ we have that $h_\delta(x) \geq \lim_{x\to\infty} g(x) \geq g(x)$ on $(1,\infty)$.

For $\beta \in (1, \infty)$, $\frac{1-\beta}{\beta} < 0$. 
It follows that $\left(1+x^{-\beta}\right)^\frac{1-\beta}{\beta} \to 0$ as $x \to 0$, and hence $\lim_{x\to 0}g(x)$ exists and is finite.
Since $g(x)$ is decreasing on $(0,1)$ and increasing on $(1,\infty)$, it follows that 
$g(x) \leq \max\{\lim_{x\to 0}g(x), \lim_{x\to \infty}g(x)\}$ for all $x\in [0, \infty)$

\end{proof}

\subsection{Proof of Theorem \ref{thm:convergence-rate-general}}\label{proof:thm2}

\begin{theorem}[Rates of the bias]
If $\E_{X\sim Q_X, Z\sim P_Z}\bigl[ q^4(Z|X) / p^4(Z) \bigr]$ is finite then
the bias $\E_{\XN}\bigl[D_f( \hat{Q}_Z^N \| P_Z)\bigr] - D_f\left( Q_{Z} \| P_Z\right)$ decays with rate as given in the second row of Table \ref{table:convergence}.
\end{theorem}

\begin{proof}
For each $f$-divergence we will work with the function $f_0$ which is decreasing on $(0,1)$ and increasing on $(1, \infty)$ with $D_f = D_{f_0}$ (see Appendix \ref{appendix:f-fns}).

For shorthand we will sometimes use the notation $\| q(z|X)/p(z)\|^2_{L_2(P_Z)} = \int \frac{q(z|X)^2}{p(z)^2} p(z) dz$ and $\| q^2(z|X)/p^2(z)\|^2_{L_2(P_Z)} = \int \frac{q(z|X)^4}{p(z)^4} p(z) dz$.

We will denote $C:= \E_{X\sim Q_X, Z\sim P_Z}\bigl[ q^4(Z|X) / p^4(Z) \bigr]$ which is finite by assumption. 
This implies that the second moment $B:= \E_{X\sim Q_X, Z\sim P_Z}\bigl[ q^2(Z|X) / p^2(Z) \bigr]$ is also finite, thanks to Jensen's inequality: 
\[
\E[Y^2] = \E[\sqrt{Y^4}]\leq \sqrt{\E[Y^4]}.
\]

\paragraph{The case that $D_f$ is the $\chi^2$-divergence:}
In this case, using $f(x) = x^2-1$, it can be seen that the bias is equal to
\begin{align}\label{eqn:chi2-bias}
    \E_{\XN} \left[ D_{f}\left( \hat{Q}_Z^N \| P_Z\right)\right] - D_{f}\left( Q_{Z} \| P_Z\right) = \E_{\XN}\left[ \int_Z  \left(\frac{\hat{q}_N(z) - q(z)}{p(z)}\right)^2 dP(z) \right].
\end{align}
Indeed, expanding the right hand side and using the fact that $\E_{\XN}\hat{q}_N(z) = q(z)$ yields
\begin{align*}
    &\E_{\XN}\left[ \int_Z  \frac{\hat{q}^2_N(z) - 2\hat{q}_N(z)q(z) + q^2(z)}{p^2(z)} dP(z) \right]\\
    &=\E_{\XN}\left[ \int_Z  \frac{\hat{q}^2_N(z) - q^2(z)}{p^2(z)} dP(z) \right]\\
    &=\E_{\XN}\left[ \int_Z  \left(\frac{\hat{q}^2_N(z)}{p^2(z)} - 1 \right)dP(z) \right] - \int_Z  \left(\frac{q^2(z)}{p^2(z)} - 1\right) dP(z)\\
    &= \E_{\XN} \left[ D_{f}\left( \hat{Q}_Z^N \| P_Z\right)\right] - D_{f}\left( Q_{Z} \| P_Z\right).
\end{align*}
Again using the fact that $\E_{\XN}\hat{q}_N(z) = q(z)$, observe that taking expectations over $\XN$ in the right hand size of Equation \ref{eqn:chi2-bias} above (after changing the order of integration) can be viewed as taking the variance of $\hat{q}_N(z)/p(z)$, the average of $N$ i.i.d. random variables, and so

\begin{align*}
 \E_{\XN}\left[ \int_Z  \left(\frac{\hat{q}_N(z) - q(z)}{p(z)}\right)^2 dP(z) \right] &= \int_Z \E_{\XN}\left[  \left(\frac{\hat{q}_N(z) - q(z)}{p(z)}\right)^2 \right] dP(z) \\
 &=\frac{1}{N} \int_Z \E_{X}\left[  \left(\frac{q(z|X) - q(z)}{p(z)}\right)^2 \right] dP(z) \\
 &= \frac{1}{N} \E_{X} \chi^2\left( Q_{Z|X} \| P_Z \right) - \frac{1}{N} \chi^2\left( Q_{Z} \| P_Z \right)\\
 &\leq \frac{B-1}{N}.
\end{align*}

\paragraph{The case that $D_f$ is the Total Variation distance or $D_{f_\beta}$ with $\beta>1$:}

For these divergences, we only need the condition that the second moment $\E_X \| q(z|X)/p(z)\|^2_{L_2(P_Z)} < \infty$ is bounded.

\begin{align*}
    & \E_{\XN} \left[ D_{f_0}\left( \hat{Q}_Z^N \| P_Z\right)\right] - D_{f_0}\left( Q_{Z} \| P_Z\right)  \\
    &= \E_{\XN}\E_{P_Z} \left[ f_0\left( \frac{\hat{q}_N(z)}{p(z)} \right) - f_0\left( \frac{q(z)}{p(z)} \right) \right] \\
    &\leq \E_{\XN}\E_{P_Z} \left[ \left( \frac{\hat{q}_N(z) - q(z)}{p(z)} \right)     f_0'\left(\frac{\hat{q}_N(z)}{p(z)} \right) \right]  \\
    &\leq \underbrace{ \sqrt{ \E_{\XN}\E_{P_Z} \left[\left( \frac{\hat{q}_N(z) - q(z)}{p(z)} \right)^2\right] }}_{(i)} 
    \times 
    \underbrace{\sqrt{\E_{\XN}\E_{P_Z} \left[ f_0'^2\left(\frac{\hat{q}_N(z)}{p(z)}\right] \right) }}_{(ii)}
\end{align*}
where the first inequality holds due to convexity of $f_0$ and the second inequality follows by Cauchy-Schwartz.
Then,
\begin{align*}
    (i)^2 
    &= \E_{P_Z} \text{Var}_{\XN}\left[\frac{\hat{q}_N(z)}{p(z)} \right]\\
    &= \frac{1}{N}\E_{P_Z} \text{Var}_{X}\left[\frac{q(z|X)}{p(z)} \right]\\
    &\leq \frac{1}{N}\E_{X} \E_{P_Z} \left[ \frac{q^2(z|X)}{p^2(z)} \right] = \frac{1}{N} \E_X \left\| \frac{q(z|X)}{p(z)} \right\|^2_{L_2(P_Z)}\\
    \implies & (i) = O\left(\frac{1}{\sqrt{N}}\right).
\end{align*}

For Total Variation, ${f'_0}^2(x) \leq 1$, so
\begin{align*}
    (ii)^2 \leq 1.
\end{align*}

For $D_{f_\beta}$ with $\beta>1$, Lemma~\ref{lemma:upper-bound-f-beta} shows that $f'^2_0(x) \leq \max\{\lim_{x\to 0}f'^2_0(x), \lim_{x\to \infty}f'^2_0(x)\} < \infty$ and so 
\begin{align*}
    (ii)^2 = O(1).
\end{align*}

Thus, for both cases considered,
\begin{align*}
    \E_{\XN} \left[ D_f\left( \hat{Q}_Z^N \| P_Z\right)\right] - D_f\left( Q_{Z} \| P_Z\right) \leq O\left( \frac{1}{\sqrt{N}} \right).
\end{align*}

\paragraph{All other divergences.}

We start by writing the difference as the sum of integrals over mutually exclusive events that partition $\mathcal{Z}$.
Denoting by $\gamma_N$ and $\delta_N$ scalars depending on $N$, write

\begin{align*}
    & \E_{\XN} \left[ D_f\left( \hat{Q}_Z^N \| P_Z\right)\right] - D_f\left( Q_{Z} \| P_Z\right)  \\
    &= \E_{\XN} \left[\int f_0\left( \frac{\hat{q}_N(z)}{p(z)} \right) - f_0\left( \frac{q(z)}{p(z)} \right) dP_Z(z) \right] \\
    &= \E_{\XN} \left[ \int f_0\left( \frac{\hat{q}_N(z)}{p(z)} \right) - f_0\left( \frac{q(z)}{p(z)} \right) \mathds{1}_{\left\lbrace \frac{\hat{q}_N(z)}{p(z)} \leq \delta_N \text{ and } \frac{q(z)}{p(z)} \leq \gamma_N \right\rbrace} dP_Z(z) \right] \tag*{\encircle{A}}\\
    & \quad + \E_{\XN} \left[\int f_0\left( \frac{\hat{q}_N(z)}{p(z)} \right) - f_0\left( \frac{q(z)}{p(z)} \right) \mathds{1}_{\left\lbrace \frac{\hat{q}_N(z)}{p(z)} \leq \delta_N \text{ and } \frac{q(z)}{p(z)} > \gamma_N \right\rbrace} dP_Z(z) \right]\tag*{\encircle{B}}\\
    & \quad + \E_{\XN} \left[ \int f_0\left( \frac{\hat{q}_N(z)}{p(z)} \right) - f_0\left( \frac{q(z)}{p(z)} \right) \mathds{1}_{\left\lbrace \frac{\hat{q}_N(z)}{p(z)} > \delta_N \right\rbrace} dP_Z(z) \right]. \tag*{\encircle{C}}
\end{align*}

Consider each of the terms \encircle{A}, \encircle{B} and  \encircle{C} separately.

Later on, we will pick $\delta_N < \gamma_N$ to be decreasing in $N$.
In the worst case, $N>8$ will be sufficient to ensure that $\gamma_N < 1$, so in the remainder of this proof we will assume that $\delta_N, \gamma_N < 1$.

\encircle{A}: 
Recall that $f_0(x)$ is decreasing on the interval $[0,1]$.
Since $\gamma_N, \delta_N \leq 1$, the integrand is at most $f_0(0) - f_0(\gamma_N)$, and so
\begin{align*}
    \encircle{A} &\leq f_0(0) - f_0(\gamma_N).
\end{align*}

\encircle{B}:
The integrand is bounded above by $f_0(0)$ since $\delta_N<1$, and so 
\begin{align*}
    \encircle{B} &\leq f_0(0) \times \mathbb{P}_{Z,\XN}
    \underbrace{
    \left\lbrace \frac{\hat{q}_N(z)}{p(z)} \leq \delta_N \text{ and } \frac{q(z)}{p(z)} > \gamma_N \right\rbrace}_{\encircle{$*$}}.
\end{align*}

We will upper bound $\mathbb{P}_{Z,\XN} \encircle{$*$}$:
observe that if $\gamma_N > \delta_N$, then  $\encircle{$*$}\implies \left| \frac{\hat{q}_N(z) - q(z)}{p(z)} \right| \geq \gamma_N - \delta_N$.
It thus follows that
\begin{align*}
    \mathbb{P}_{Z,\XN} \encircle{$*$} &\leq  \mathbb{P}_{Z,\XN} \left\lbrace \left| \frac{\hat{q}_N(z) - q(z)}{p(z)} \right| \geq \gamma_N - \delta_N  \right\rbrace\\
    &= \mathbb{E}_Z\left[ \mathbb{P}_{\XN} \left\lbrace \left| \frac{\hat{q}_N(z) - q(z)}{p(z)} \right| \geq \gamma_N - \delta_N  \  | \ Z \right\rbrace \right]\\
    &\leq \mathbb{E}_Z\left[ 
    \frac{\text{Var}_{\XN}\left[ 
    \frac{\hat{q}_N(z)}{p(z)}
    \right]}{(\gamma_N - \delta_N)^2}
    \right]\\
    &= \frac{1}{N(\gamma_N - \delta_N)^2}  \E_Z \left[ \E_{X}\left[
     \frac{q^2(z|X)}{p^2(z)}\right] - \frac{q^2(z)}{p^2(z)} \right]\\
    &\leq \frac{1}{N(\gamma_N - \delta_N)^2}  \E_Z \E_{X}\left[
     \frac{q^2(z|X)}{p^2(z)}\right]\\
    &\leq \frac{\sqrt{C} }{N(\gamma_N - \delta_N)^2}.\\
\end{align*}
The second inequality follows by Chebyshev's inequality, noting that $\E_{\XN} \frac{\hat{q}_N(z)}{p(z)} = \frac{q(z)}{p(z)}$.
The penultimate inequality is due to dropping a negative term.
The final inequality is due to the boundedness assumption $C =  \E_{X}\left\| \frac{q^2(z|X)}{p^2(z)} \right\|^2_{L_2(P_Z)}$.
We thus have that 
\begin{align*}
    \encircle{B}
    &\leq f_0(0) \frac{\sqrt{C}}{N (\gamma_N - \delta_N)^2}.
\end{align*}

\encircle{C}:
Bounding this term will involve two computations, one of which $(\dagger\dagger)$ will be treated separately for each divergence we consider.

\begin{align*}
    \encircle{C} &= \E_{\XN}\left[\int f_0\left( \frac{\hat{q}_N(z)}{p(z)}\right) - f_0\left( \frac{q(z)}{p(z)} \right) \mathds{1}_{\left\lbrace \frac{\hat{q}_N(z)}{p(z)} > \delta_N \right\rbrace} dP_Z(z)\right] \\
    &\leq \E_{\XN} \left[ \int \left(\frac{\hat{q}_N(z)}{p(z)} - \frac{q(z)}{p(z)}\right) f'_0\left(\frac{\hat{q}_N(z)}{p(z)} \right) \mathds{1}_{\left\lbrace \frac{\hat{q}_N(z)}{p(z)} > \delta_N \right\rbrace} dP_Z(z) \right]
    && \text{(Convexity of $f$)}
    \\
    &\leq \underbrace{\sqrt{\E_{\XN} \E_{Z}\left[ \left(\frac{\hat{q}_N(z)}{p(z)} - \frac{q(z)}{p(z)}\right)^2 \right]}}_{(\dagger)} \times 
    \underbrace{\sqrt{\E_{\XN} \E_{Z} \left[ f'^2_0\left(\frac{\hat{q}_N(z)}{p(z)} \right) \mathds{1}_{\left\lbrace \frac{\hat{q}_N(z)}{p(z)} > \delta_N \right\rbrace} \right]}}_{(\dagger\dagger)}
    && \text{(Cauchy-Schwartz)}
\end{align*}
Noting that $\E_{X} \frac{q(z|X)}{p(z)} = \frac{q(z)}{p(z)}$, we have that
\begin{align*}
    (\dagger)^2
    &= \E_Z \text{Var}_{\XN} \left[ \frac{\hat{q}_N(z)}{p(z)}\right] \\
    &=  \frac{1}{N} \E_Z \text{Var}_{X} \left[ \frac{q(z|X)}{p(z)}\right] \\
    &\leq \frac{1}{N} \E_X \left\|\frac{q(z|X)}{p(z)} \right\|^2_{L_2(P_Z)}\\
    \implies (\dagger) &\leq \frac{\sqrt{B}}{\sqrt{N}}
\end{align*}
where $\sqrt{B} = \sqrt{\E_X \left\|\frac{q(z|X)}{p(z)} \right\|^2_{L_2(P_Z)}}$ is finite by assumption.

Term $(\dagger\dagger)$ will be bounded differently for each divergence, though using a similar pattern. 
The idea is to use the results of Lemmas~\ref{lemma:concave-upper-bound-kl}-\ref{lemma:upper-bound-f-beta} in order to upper bound $f'^2_0(x)$ with something that can be easily integrated.

\paragraph{KL.}

By Lemma \ref{lemma:concave-upper-bound-kl}, there exists a function $h_{\delta_N}(x)$ that is positive and concave on $[0, \infty)$ and is an upper bound of $f_0'^2(x)$ on $[\delta_N, \infty)$ with $h_{\delta_N}(1) = \log^2(\delta_N) + \frac{2}{e}$.

\begin{align*}
    (\dagger\dagger)^2 
    &= \E_{\XN} \left[\int f_0'^2\left(\frac{\hat{q}_N(z)}{p(z)} \right) \mathds{1}_{\left\lbrace \frac{\hat{q}_N(z)}{p(z)} > \delta_N \right\rbrace} p(z) dz \right]
    \\
    &\leq \E_{\XN} \left[ \int h_{\delta_N}\left(\frac{\hat{q}_N(z)}{p(z)} \right) \mathds{1}_{\left\lbrace \frac{\hat{q}_N(z)}{p(z)} > \delta_N \right\rbrace} p(z) dz \right]
    && \text{($h_{\delta_N}$ upper bounds $f'^2$ on $(\delta_N, \infty)$)}
    \\
    &\leq \E_{\XN}\left[ \int h_{\delta_N}\left(\frac{\hat{q}_N(z)}{p(z)} \right) p(z) dz \right]
    && \text{($h_{\delta_N}$ non-negative on $[0, \infty)$)}
    \\
    &\leq \E_{\XN} \left[ h_{\delta_N}\left( \int \frac{\hat{q}_N(z)}{p(z)} p(z) dz \right)  \right]
    && \text{($h_{\delta_N}$ concave)}
    \\
    &=  h_{\delta_N}\left( 1\right) \\
    &= \log^2(\delta_N) + \frac{2}{e} \\
    \implies (\dagger\dagger)&= \sqrt{\log^2(\delta_N) + \frac{2}{e}}.
\end{align*}

Therefore,
\begin{align*}
    \encircle{C} \leq \sqrt{B} \sqrt{\frac{\log^2(\delta_N) + \frac{2}{e}}{N}}.
\end{align*}

Putting everything together,

\begin{align*}
    &\E_{\XN} \left[ D_f\left( \hat{Q}_Z^N \| P_Z\right)\right] - D_f\left( Q_{Z} \| P_Z\right)\\
    &\leq \encircle{A} + \encircle{B} + \encircle{C} \\
    &\leq f_0(0) - f_0(\gamma_N) + f_0(0) \frac{\sqrt{C}}{N \left( \gamma_N - \delta_N \right)^2} + \sqrt{B} \sqrt{\frac{\log^2(\delta_N) + \frac{2}{e}}{N}}\\
    &= \gamma_N - \gamma_N \log \gamma_N  + \frac{\sqrt{C}}{N \left( \gamma_N - \delta_N \right)^2} + \sqrt{B} \sqrt{\frac{\log^2(\delta_N) + \frac{2}{e}}{N}}.
\end{align*}
Taking $\delta_N = \frac{1}{N^{1/3}}$ and $\gamma_N = \frac{2}{N^{1/3}}$:
\begin{align*}
    &=\frac{2}{N^{1/3}} - \frac{2}{N^{1/3}} \log\left( \frac{2}{N^{1/3}}\right) + \frac{ \sqrt{C} }{N \cdot \frac{1}{N^{2/3}} } + \sqrt{B} \sqrt{\frac{\log^2\left(\frac{1}{N^{1/3}}\right) + \frac{2}{e}}{N}}\\
    &= \frac{ 2 - 2\log2}{N^{1/3}} + \frac{2}{3}\frac{\log N}{N^{1/3}} + \frac{\sqrt{C}}{N^{1/3}} + \sqrt{B} \sqrt{\frac{\frac{1}{4}\log^2\left(N\right) + \frac{2}{e}}{N}} \\
    & = O\left( \frac{\log N}{N^{1/3}}\right)
\end{align*}

\paragraph{Squared-Hellinger.}
Lemma~\ref{lemma:upper-bound-hellinger} provides a function $h_\delta$ that upper bounds $f'^2(x)$ for $x \in \in[\delta, \infty)$.

\begin{align*}
    (\dagger\dagger)^2 
    &= \E_{\XN} \left[\int f_0'^2\left(\frac{\hat{q}_N(z)}{p(z)} \right) \mathds{1}_{\left\lbrace \frac{\hat{q}_N(z)}{p(z)} > \delta_N \right\rbrace} p(z) dz \right]
    \\
    &\leq \E_{\XN} \left[\int h_{\delta_N}\left(\frac{\hat{q}_N(z)}{p(z)} \right) \mathds{1}_{\left\lbrace \frac{\hat{q}_N(z)}{p(z)} > \delta_N \right\rbrace} p(z) dz \right]
    && \text{($h_{\delta_N}$ upper bounds $f_0'^2$ on $(\delta_N, \infty)$)}
    \\
    &\leq \E_{\XN} \left[\int h_{\delta_N}\left(\frac{\hat{q}_N(z)}{p(z)} \right) p(z) dz \right]
    && \text{($h_{\delta_N}$ non-negative on $[0, \infty)$)}
    \\
    &= \frac{1}{\delta_N} \E_{\XN}\E_{P_Z} \left[ \left(\frac{\hat{q}_N(z)}{p(z)} - 1 \right)^2 \right] \\
    &\leq \frac{1}{\delta_N} \E_{\XN}\E_{P_Z} \left[ \left(\frac{\hat{q}_N(z)}{p(z)}\right)^2 + 1 \right] \\
    &= \frac{1}{\delta_N} + \frac{1}{\delta_N} \E_{\XN}\left[ \left\| \frac{\hat{q}_N(z)}{p(z)}\right\|^2_{L_2(P_Z)} \right]\\
    &\leq \frac{B + 1}{\delta_N} \\
    \implies (\dagger\dagger)&= \frac{\sqrt{B + 1}}{\sqrt{\delta_N}}.
\end{align*}

and thus

\begin{align*}
    &\E_{\XN} \left[ D_f\left( \hat{Q}_Z^N \| P_Z\right)\right] - D_f\left( Q_{Z} \| P_Z\right)\\
    &\leq \encircle{A} + \encircle{B} + \encircle{C} \\
    &\leq f_0(0) - f_0(\gamma_N) + f_0(0) \frac{\sqrt{C}}{N \left( \gamma_N - \delta_N \right)^2} + \frac{\sqrt{B}\sqrt{B + 1}}{\sqrt{N \delta_N}}\\
    &= 2\sqrt{\gamma_N}  + \frac{2\sqrt{C}}{N \left( \gamma_N - \delta_N \right)^2} + \frac{\sqrt{B}\sqrt{B + 1}}{\sqrt{N \delta_N}}.
\end{align*}

Setting $\gamma_N = \frac{2}{N^{2/5}}$ and $\delta_N = \frac{1}{N^{2/5}}$ yields

\begin{align*}
    &= \frac{2}{N^{1/5}}  + \frac{2\sqrt{C}}{N^{1/5}} + \frac{\sqrt{B}\sqrt{B + 1}}{N^{3/10}} \\
    & = O\left(\frac{1}{N^{1/5}} \right)
\end{align*}

\paragraph{$\alpha$-divergence with $\alpha\in(-1,1)$.}
Lemma~\ref{lemma:upper-bound-alpha} provides a function $h_\delta$ that upper bounds $f'^2(x)$ for $x \in \in[\delta, \infty)$.

\begin{align*}
    (\dagger\dagger)^2 
    &= \E_{\XN} \left[\int f_0'^2\left(\frac{\hat{q}_N(z)}{p(z)} \right) \mathds{1}_{\left\lbrace \frac{\hat{q}_N(z)}{p(z)} > \delta_N \right\rbrace} p(z) dz \right]
    \\
    &\leq \E_{\XN} \left[\int h_{\delta_N}\left(\frac{\hat{q}_N(z)}{p(z)} \right) \mathds{1}_{\left\lbrace \frac{\hat{q}_N(z)}{p(z)} > \delta_N \right\rbrace} p(z) dz \right]
    && \text{($h_{\delta_N}$ upper bounds $f_0'^2$ on $(\delta_N, \infty)$)}
    \\
    &\leq \E_{\XN} \left[\int h_{\delta_N}\left(\frac{\hat{q}_N(z)}{p(z)} \right) p(z) dz \right]
    && \text{($h_{\delta_N}$ non-negative on $[0, \infty)$)}
    \\
    &= \frac{4\left(\delta_N^\frac{\alpha-1}{2} - 1\right)^2}{(\alpha - 1)^2 (\delta_N- 1)^2} \E_{\XN}\E_{P_Z} \left[ \left(\frac{\hat{q}_N(z)}{p(z)} - 1 \right)^2 \right] \\
    &\leq \frac{4\left(\delta_N^\frac{\alpha-1}{2} - 1\right)^2}{(\alpha - 1)^2 (\delta_N- 1)^2} \E_{\XN}\E_{P_Z} \left[ \left(\frac{\hat{q}_N(z)}{p(z)}\right)^2 + 1 \right] \\
    &= \frac{4\left(\delta_N^\frac{\alpha-1}{2} - 1\right)^2}{(\alpha - 1)^2 (\delta_N- 1)^2}\left( 1 +  \E_{\XN}\left[ \left\| \frac{\hat{q}_N(z)}{p(z)}\right\|^2_{L_2(P_Z)} \right] \right)\\
    &\leq \frac{4(1+B)\left(\delta_N^\frac{\alpha-1}{2} - 1\right)^2}{(\alpha - 1)^2 (\delta_N- 1)^2} \\
    \implies (\dagger\dagger)&= \frac{2\sqrt{1+B}\left(\delta_N^\frac{\alpha-1}{2} - 1\right)}{(\alpha - 1) (\delta_N- 1)}.
\end{align*}

and thus

\begin{align*}
    &\E_{\XN} \left[ D_f\left( \hat{Q}_Z^N \| P_Z\right)\right] - D_f\left( Q_{Z} \| P_Z\right)\\
    &\leq \encircle{A} + \encircle{B} + \encircle{C} \\
    &\leq f_0(0) - f_0(\gamma_N) + f_0(0) \frac{\sqrt{C}}{N \left( \gamma_N - \delta_N \right)^2} + \frac{2\sqrt{B}\sqrt{1+B}\left(\delta_N^\frac{\alpha-1}{2} - 1\right)}{(\alpha - 1) (\delta_N- 1) \sqrt{N}}\\
    &\leq k_1 \gamma_N^{\frac{\alpha+1}{2}} + k_2 \gamma_N + \frac{k_3}{N(\gamma_N - \delta_N)^2} + \frac{k_4 \delta_N^\frac{\alpha-1}{2}}{\sqrt{N}}.
\end{align*}
where each $k_i$ is a positive constant independent of $N$.

Setting $\gamma_N = \frac{2}{N^\frac{2}{\alpha+5}}$ and $\delta_N = \frac{1}{N^\frac{2}{\alpha+5}}$ yields

\begin{align*}
    &= \leq  \frac{k_1}{N^{\frac{\alpha+1}{\alpha+5}}} + \frac{k_2}{N^{\frac{2}{\alpha+5}}}
    + \frac{k_3}{N^{\frac{\alpha+1}{\alpha+5}}} 
    + \frac{k_4}{N^{\frac{7-\alpha}{2(\alpha+5)}}} \\
    & = O\left(\frac{1}{N^\frac{\alpha+1}{\alpha+5}} \right)
\end{align*}

\paragraph{Jensen-Shannon.}
Lemma~\ref{lemma:upper-bound-JS} provides a function $h_\delta$ that upper bounds $f'^2(x)$ for $x \in[\delta, \infty)$.

\begin{align*}
    (\dagger\dagger)^2 
    &= \E_{\XN} \left[\int f_0'^2\left(\frac{\hat{q}_N(z)}{p(z)} \right) \mathds{1}_{\left\lbrace \frac{\hat{q}_N(z)}{p(z)} > \delta_N \right\rbrace} p(z) dz \right]
    \\
    &\leq \E_{\XN} \left[\int h_{\delta_N}\left(\frac{\hat{q}_N(z)}{p(z)} \right) \mathds{1}_{\left\lbrace \frac{\hat{q}_N(z)}{p(z)} > \delta_N \right\rbrace} p(z) dz \right]
    && \text{($h_{\delta_N}$ upper bounds $f_0'^2$ on $(\delta_N, \infty)$)}
    \\
    &\leq \E_{\XN} \left[\int h_{\delta_N}\left(\frac{\hat{q}_N(z)}{p(z)} \right) p(z) dz \right]
    && \text{($h_{\delta_N}$ non-negative on $[0, \infty)$)}
    \\
    &= 5\log^2 2 + \log^2\left( \frac{\delta_N}{1+\delta_N} \right) + 2\log 2 \log\left( \frac{\delta_N}{1+\delta_N} \right)\\
    &= 5\log^2 2 + \log^2\left(1 + \frac{1}{\delta_N}\right) - 2\log 2 \log\left(1 + \frac{1}{\delta_N}\right)\\
    &\leq 5\log^2 2 + 5\log^2\left(1 + \frac{1}{\delta_N}\right) + 10\log 2 \log\left(1 + \frac{1}{\delta_N}\right)\\
    &=5\left(\log\left(1 + \frac{1}{\delta_N} \right) - \log 2\right)^2\\
    \implies (\dagger\dagger)&\leq 
    \sqrt{5}\log\left(1 + \frac{1}{\delta_N} \right) - \sqrt{5}\log 2 \\
    &\leq \sqrt{5}\log\left(\frac{2}{\delta_N} \right) - \sqrt{5}\log 2 && \text{(since $\delta_N<1$)}\\
    &= -\sqrt{5}\log(\delta_N).
\end{align*}

and thus

\begin{align*}
    &\E_{\XN} \left[ D_f\left( \hat{Q}_Z^N \| P_Z\right)\right] - D_f\left( Q_{Z} \| P_Z\right)\\
    &\leq \encircle{A} + \encircle{B} + \encircle{C} \\
    &\leq f_0(0) - f_0(\gamma_N) + f_0(0) \frac{\sqrt{C}}{N \left( \gamma_N - \delta_N \right)^2} - \frac{\sqrt{5}\sqrt{B}\log \delta_N}{\sqrt{N}}\\
    &\leq \gamma_N \log\left(\frac{1+\gamma_N}{2\gamma_N}\right) + \log(1+\gamma_N) + \frac{\log2\sqrt{C}}{N \left( \gamma_N - \delta_N \right)^2} - \frac{\sqrt{5}\sqrt{B} \log \delta_N}{\sqrt{N}} \\
\end{align*}
Using the fact that $\gamma_N \log(1+\gamma_N) \leq \gamma_N \log2$ for $\gamma_N < 1$ and $\log(1+ \gamma_N) \leq \gamma_N$, we can upper bound the last line with
\begin{align*}
    &\leq \gamma_N \left(\log2 + 1\right)   - \gamma_N \log \gamma_N  + \frac{\log2\sqrt{C}}{N \left( \gamma_N - \delta_N \right)^2} - \frac{\sqrt{5}\sqrt{B} \log \delta_N}{\sqrt{N}} \\
\end{align*}

Setting $\gamma_N = \frac{2}{N^\frac{1}{3}}$ and $\delta_N = \frac{1}{N^\frac{1}{3}}$ yields

\begin{align*}
    &= \frac{k_1}{N^{\frac{1}{3}}} + \frac{k_2 \log N}{N^{\frac{1}{3}}}
    + \frac{k_3}{N^{\frac{1}{3}}} 
    + \frac{k_4 \log N}{N^{\frac{1}{2}}} \\
    & = O\left(\frac{\log N}{N^\frac{1}{3}} \right)
\end{align*}
where the $k_i$ are positive constants independent of $N$.

\paragraph{$f_\beta$-divergence with $\beta\in(\frac{1}{2}, 1)$.}
Lemma~\ref{lemma:upper-bound-f-beta} provides a function $h_\delta$ that upper bounds $f'^2(x)$ for $x \in[\delta, \infty)$.

\begin{align*}
    (\dagger\dagger)^2 
    &= \E_{\XN} \left[ \int f_0'^2\left(\frac{\hat{q}_N(z)}{p(z)} \right) \mathds{1}_{\left\lbrace \frac{\hat{q}_N(z)}{p(z)} > \delta_N \right\rbrace} p(z) dz \right]
    \\
    &\leq \E_{\XN} \left[\int h_{\delta_N}\left(\frac{\hat{q}_N(z)}{p(z)} \right) \mathds{1}_{\left\lbrace \frac{\hat{q}_N(z)}{p(z)} > \delta_N \right\rbrace} p(z) dz \right]
    \qquad \text{($h_{\delta_N}$ upper bounds $f_0'^2$ on $(\delta_N, \infty)$)}
    \\
    &\leq \E_{\XN} \left[\int h_{\delta_N}\left(\frac{\hat{q}_N(z)}{p(z)} \right) p(z) dz \right]
    \qquad \text{($h_{\delta_N}$ non-negative on $[0, \infty)$)}
    \\
    &= \left(\frac{\beta}{1-\beta}\right)^2\left[ \left(1+\delta_N^{-\beta}\right)^\frac{1-\beta}{\beta}  - 2^{\frac{1-\beta}{\beta}}\right]^2 + \frac{\beta^2}{(1-\beta)^2}\left(2^{\frac{1-\beta}{\beta}}\right)^2\\
    &\leq 2\left(\frac{\beta}{1-\beta}\right)^2\left[ \left(1+\delta_N^{-\beta}\right)^\frac{1-\beta}{\beta}  + 2^{\frac{1-\beta}{\beta}}\right]^2\\
    &\leq 2\left(\frac{\beta}{1-\beta}\right)^2\left[ 2\left(2\delta_N^{-\beta}\right)^\frac{1-\beta}{\beta}\right]^2 
    \qquad (\text{since $\delta_N<1$ and $\beta>0$ implies $\delta_N^{-\beta} > 1$})\\
    &= 2^{\frac{2+\beta}{\beta}}\left(\frac{\beta}{1-\beta}\right)^2 \delta_N^{2(\beta - 1)}\\
    \implies (\dagger\dagger)&\leq 
    2^{\frac{2+\beta}{2\beta}}\left(\frac{\beta}{1-\beta}\right) \delta_N^{\beta - 1}\\
\end{align*}

(noting that $\frac{\beta^2}{(1-\beta)^2}\left(2^{\frac{1}{\beta} - 1}\right)^2 = \lim_{x\to\infty} f'^2_0(x)$ as defined in Lemma~\ref{lemma:upper-bound-f-beta}). Thus

\begin{align*}
    &\E_{\XN} \left[ D_f\left( \hat{Q}_Z^N \| P_Z\right)\right] - D_f\left( Q_{Z} \| P_Z\right)\\
    &\leq \encircle{A} + \encircle{B} + \encircle{C} \\
    &\leq f_0(0) - f_0(\gamma_N) + f_0(0) \frac{\sqrt{C}}{N \left( \gamma_N - \delta_N \right)^2} + \frac{\sqrt{B}}{\sqrt{N}}2^{\frac{2+\beta}{2\beta}}\left(\frac{\beta}{1-\beta}\right) \delta_N^{\beta - 1}\\
    &\leq \frac{\beta}{1-\beta}\left[1 - \left(1+\delta_N^\beta\right)^{1/\beta} + 2^{\frac{1-\beta}{\beta}}\delta_N\right] + f_0(0) \frac{\sqrt{C}}{N \left( \gamma_N - \delta_N \right)^2} + \frac{\sqrt{B}}{\sqrt{N}}2^{\frac{2+\beta}{2\beta}}\left(\frac{\beta}{1-\beta}\right) \delta_N^{\beta - 1}\\
    &\leq \frac{\beta}{1-\beta} 2^{\frac{1-\beta}{\beta}}\delta_N + f_0(0) \frac{\sqrt{C}}{N \left( \gamma_N - \delta_N \right)^2}  + \frac{\sqrt{B}}{\sqrt{N}}2^{\frac{2+\beta}{2\beta}}\left(\frac{\beta}{1-\beta}\right) \delta_N^{\beta - 1}\\
    &= k_1 \delta_N + \frac{k_2}{N(\gamma_N - \delta_N)^2} + \frac{k_3\delta_N^{\beta-1}}{\sqrt{N}}
\end{align*}
where the $k_i$ are positive constants independent of $N$.

Setting $\gamma_N = \frac{2}{N^\frac{1}{3}}$ and $\delta_N = \frac{1}{N^\frac{1}{3}}$ yields

\begin{align*}
    &= \frac{k_1}{N^{\frac{1}{3}}}
    + \frac{k_2}{N^{\frac{1}{3}}} 
    + \frac{k_3}{N^{\frac{1}{2}+\frac{\beta-1}{3}}} \\
    &= O\left(\frac{1}{N^\frac{1}{3}}\right)
\end{align*}

\end{proof}
\subsection{Proof of Theorem \ref{thm:concentration}}\label{proof:thm3}

We will make use of McDiarmid's theorem in our proof of Theorem \ref{thm:concentration}:

\begin{theorem*}[McDiarmid's inequality]
Suppose that $X_1, \ldots, X_N \in \mathcal{X}$ are independent random variables and that $\phi : \mathcal{X}^N \to \R$ is a function. 
If it holds that for all $i\in\{1,\ldots,N\}$ and $x_1, \ldots, x_N, x_{i'}$, 
\begin{align*}
    \left| \phi(x_1, \ldots, x_{i-1}, x_i, x_{i+1}, \ldots, x_N) - \phi(x_1, \ldots, x_{i-1}, x_{i'}, x_{i+1}, \ldots, x_N)\right| \leq c_i,
\end{align*}
then
\begin{align*}
    \mathbb{P} \left(\phi(X_1,\ldots, X_N) - \E\phi \geq t \right) \leq \exp\left(\frac{-2t^2}{\sum_{i=1}^N c^2_i} \right)
\end{align*}
and
\begin{align*}
    \mathbb{P} \left(\phi(X_1,\ldots, X_N) - \E\phi \geq -t \right) \leq \exp\left(\frac{-2t^2}{\sum_{i=1}^N c^2_i} \right)
\end{align*}
\end{theorem*}

In our setting we will consider $\phi(\XN) = D_f\left( \hat{Q}_Z^N \| P_Z\right)$.

\begin{theorem}[Tail bounds for RAM]
Suppose that ${\chi^2\left(Q_{Z|x} \| P_Z\right) \leq C < \infty}$ for all $x$ and for some constant $C$.
Then, the RAM estimator ${D_f( \hat{Q}_Z^N \| P_Z)}$ concentrates to its mean in the following sense. 
For $N>8$ and for any $\delta >0$, with probability at least $1-\delta$ it holds that
\begin{align*}
    \left| D_f( \hat{Q}_Z^N \| P_Z) - \mathbb{E}_{\XN} \bigl[D_f(\hat{Q}_Z^N \| P_Z)\bigr] \right| \leq {K \cdot \psi(N)} \  \sqrt{\log (2/\delta)},
\end{align*}
where $K$ is a constant and $\psi(N)$ is given in Table~\ref{table:concentration}.
\end{theorem}
\begin{proof}[Proof (Theorem \ref{thm:concentration})]
We will show that $D_f\left( \hat{Q}_Z^N \| P_Z\right)$ exhibits the bounded difference property as in the statement of McDiarmid's theorem.
Since $\hat{q}_N(z)$ is symmetric in the indices of $\XN$, we can without loss of generality consider only the case $i=1$.
Henceforth, suppose $\XN, {\XN}'$ are two batches of data with $\XN_1 \not= {\XN}'_1$ and $\XN_i = {\XN}'_i$ for all $i > 1$. 
For the remainder of this proof we will write explicitly the dependence of $\hat{Q}_Z^N$ on $\XN$. 
We will write $\hat{Q}_Z^N(\XN)$ for the probability measure and $\hat{q}_N(z; \XN)$ for its density.

We will show that $\left|D_f\left( \hat{Q}_Z^N(\XN) \| P_Z\right) - D_f\left( \hat{Q}_Z^N({\XN}') \| P_Z\right)\right| \leq c_N$ where $c_N$ is a constant depending only on $N$.
From this fact, McDiarmid's theorem and the union bound, it follows that:
\begin{align*}
    &\mathbb{P}\left( \left| D_f\left( \hat{Q}_Z^N(\XN) \| P_Z\right) - \E_{\XN} D_f\left( \hat{Q}_Z^N(\XN) \| P_Z\right) \right|\geq t \right) \\
    &= \mathbb{P}\bigg( D_f\left( \hat{Q}_Z^N(\XN) \| P_Z\right) - \E_{\XN} D_f\left( \hat{Q}_Z^N(\XN) \| P_Z\right) \geq t \text{ or }\\
    & \qquad \qquad D_f\left( \hat{Q}_Z^N(\XN) \| P_Z\right) - \E_{\XN} D_f\left( \hat{Q}_Z^N(\XN) \| P_Z\right) \leq -t\bigg) \\
    &\leq \mathbb{P}\left( D_f\left( \hat{Q}_Z^N(\XN) \| P_Z\right) - \E_{\XN} D_f\left( \hat{Q}_Z^N(\XN) \| P_Z\right) \geq t \right) 
    + \\
    & \qquad \qquad \mathbb{P}\left( D_f\left( \hat{Q}_Z^N(\XN) \| P_Z\right) - \E_{\XN} D_f\left( \hat{Q}_Z^N(\XN) \| P_Z\right) \leq -t\right) \\
    &\leq 2 \exp\left(\frac{-2 t^2 }{Nc^2_N} \right).
\end{align*}
Observe that by setting $t = \sqrt{\frac{Nc_N^2}{2} \log\left(\frac{2}{\delta}\right)}$, 

the above inequality is equivalent to the statement that for any $\delta>0$, with probability at least $1-\delta$ 
\begin{align*}
    \left| D_f\left( \hat{Q}_Z^N(\XN) \| P_Z\right) - \E_{\XN} D_f\left( \hat{Q}_Z^N(\XN) \| P_Z\right) \right| < \sqrt{\frac{Nc_N^2}{2}} \sqrt{\log\left(\frac{2}{\delta}\right)}.
\end{align*}
We will show that $c_N \leq k N^{-1/2} \psi(N)$ for $k$ and $\psi(N)$ depending on $f$.
The statement of Theorem~\ref{thm:concentration} is of this form.
Note that in order to show that
\begin{align}\label{eqn:bounded-diff-abs}
    \left|D_f\left( \hat{Q}_Z^N(\XN) \| P_Z\right) - D_f\left( \hat{Q}_Z^N({\XN}') \| P_Z\right)\right| \leq c_N,
\end{align}
it is sufficient to prove that 
\begin{align}\label{eqn:bounded-diff}
     D_f\left( \hat{Q}_Z^N(\XN) \| P_Z\right) - D_f\left( \hat{Q}_Z^N({\XN}') \| P_Z\right) \leq c_N
\end{align}
since the symmetry in $\XN \leftrightarrow {\XN}'$ implies that
\begin{align}
    - D_f\left( \hat{Q}_Z^N(\XN) \| P_Z\right) + D_f\left( \hat{Q}_Z^N({\XN}') \| P_Z\right) \leq c_N
\end{align}
and thus implies Inequality \ref{eqn:bounded-diff-abs}.
The remainder of this proof is therefore devoted to showing that Inequality \ref{eqn:bounded-diff} holds for each divergence.

We will make use of the fact that $\chi^2\left(Q_{Z|x} \| P_Z\right) \leq C \implies \bigl\| \frac{q(z|x)}{p(z)} \bigr\|_{L_2(P_Z)} \leq C+1 $

\paragraph{The case that $D_f$ is the $\chi^2$-divergence, Total Variation or $D_{f_\beta}$ with $\beta>1$:}
\begin{align*}
    & D_f\left( \hat{Q}_Z^N(\XN) \| P_Z\right) - D_f\left( \hat{Q}_Z^N({\XN}') \| P_Z\right)  \\
    &= \int f_0\left( \frac{d\hat{Q}_Z^N(\XN)}{dP_Z}(z) \right) - f_0\left( \frac{d\hat{Q}_Z^N({\XN}')}{dP_Z}(z) \right) dP_Z(z)  \\
    &\leq \int \left( \frac{\hat{q}_N(z;\XN) - \hat{q}_N(z;{\XN}')}{p(z)} \right)     f'_0\left(\frac{\hat{q}_N(z;\XN)}{p(z)} \right) dP_Z(z)  \\
    &\leq \left\| \frac{\hat{q}_N(z;\XN) - \hat{q}_N(z;{\XN}')}{p(z)} \right\|_{L_2(P_Z)} \times \left\| f'_0\left(\frac{\hat{q}_N(z;\XN)}{p(z)}\right) \right\|_{L_2(P_Z)}
    && \text{ (Cauchy-Schwartz)}\\
    &= \left\| \frac{1}{N} \frac{q(z|X_1) - q(z|X'_1)}{p(z)} \right\|_{L_2(P_Z)} \times \left\| f'_0\left(\frac{\hat{q}_N(z;\XN)}{p(z)}\right) \right\|_{L_2(P_Z)} \\
    &\leq \frac{1}{N} \left( \left\| \frac{q(z|X_1)}{p(z)}\right\|_{L_2(P_Z)} + \left\|\frac{q(z|X'_1)}{p(z)} \right\|_{L_2(P_Z)} \right)\times \left\| f'_0\left(\frac{\hat{q}_N(z;\XN)}{p(z)}\right) \right\|_{L_2(P_Z)} \\
    &\leq \frac{2(C+1)}{N} \left\| f'_0\left(\frac{\hat{q}_N(z;\XN)}{p(z)}\right) \right\|_{L_2(P_Z)}.
\end{align*}

By similar arguments as made in the proof of Theorem \ref{thm:convergence-rate-general} considering the term $(ii)$, $\left\| f_0'\left(\frac{\hat{q}_N(z;\XN)}{p(z)}\right) \right\|_{L_2(P_Z)} = \sqrt{\E_Z {f'_0}^2\left(\frac{\hat{q}_N(z;\XN)}{p(z)}\right)} = O(1)$ thus we have the difference is upper-bounded by $c_N = \frac{k}{N}$ for some constant $k$.
The only modification needed to the proof in Theorem \ref{thm:convergence-rate-general} is the omission of all occurrences of $\E_{\XN}$.

This holds for any $N>0$.

\paragraph{All other divergences.}

Similar to the proof of Theorem \ref{thm:convergence-rate-general},
we write the difference as the sum of integrals over different mutually exclusive events that partition $\mathcal{Z}$.
Denoting by $\gamma_N$ and $\delta_N$ scalars depending on $N$, we have that

\begin{align*}
    & D_f\left( \hat{Q}_Z^N(\XN) \| P_Z\right) - D_f\left( \hat{Q}_Z^N({\XN}') \| P_Z\right)  \\
    &= \int f_0\left( \frac{d\hat{Q}_Z^N(\XN)}{dP_Z}(z) \right) - f_0\left( \frac{d\hat{Q}_Z^N({\XN}')}{dP_Z}(z) \right) dP_Z(z)  \\
    &= \int f_0\left( \frac{d\hat{Q}_Z^N(\XN)}{dP_Z}(z) \right) - f_0\left( \frac{d\hat{Q}_Z^N({\XN}')}{dP_Z}(z) \right) \mathds{1}_{\left\lbrace \frac{d\hat{Q}_Z^N(\XN)}{dP_Z}(z) \leq \delta_N \text{ and } \frac{d\hat{Q}_Z^N({\XN}')}{dP_Z}(z) \leq \gamma_N \right\rbrace} dP_Z(z) \tag*{\encircle{A}}\\
    & \quad + \int f_0\left( \frac{d\hat{Q}_Z^N(\XN)}{dP_Z}(z) \right) - f_0\left( \frac{d\hat{Q}_Z^N({\XN}')}{dP_Z}(z) \right) \mathds{1}_{\left\lbrace \frac{d\hat{Q}_Z^N(\XN)}{dP_Z}(z) \leq \delta_N \text{ and } \frac{d\hat{Q}_Z^N({\XN}')}{dP_Z}(z) > \gamma_N \right\rbrace} dP_Z(z) \tag*{\encircle{B}}\\
    & \quad + \int f_0\left( \frac{d\hat{Q}_Z^N(\XN)}{dP_Z}(z) \right) - f_0\left( \frac{d\hat{Q}_Z^N({\XN}')}{dP_Z}(z) \right) \mathds{1}_{\left\lbrace \frac{d\hat{Q}_Z^N(\XN)}{dP_Z}(z) > \delta_N \right\rbrace} dP_Z(z). \tag*{\encircle{C}}
\end{align*}

We will consider each of the terms \encircle{A}, \encircle{B} and  \encircle{C} separately.

Later on, we will pick $\gamma_N$ and $\delta_N$ to be decreasing in $N$ such that $\delta_N < \gamma_N$.
We will require $N$ sufficiently large so that $\gamma_N< 1$, so in the rest of this proof we will assume this to be the case and later on provide lower bounds on how large $N$ must be to ensure this.

\encircle{A}: 
Recall that $f_0(x)$ is decreasing on the interval $[0,1]$.
Since $\gamma_N, \delta_N \leq 1$, 
the integrand is at most $f_0(0) - f_0(\gamma_N)$, and so 
\begin{align*}
    \encircle{A} &\leq f_0(0) - f_0(\gamma_N)
\end{align*}

\encircle{B}:
Since $\delta_N \leq 1$,
the integrand is at most $f_0(0)$ and so
\begin{align*}
    \encircle{B} &\leq f_0(0) \times \mathbb{P}_Z
    \underbrace{
    \left\lbrace \frac{d\hat{Q}_Z^N(\XN)}{dP_Z}(z) \leq \delta_N \text{ and } \frac{d\hat{Q}_Z^N({\XN}')}{dP_Z}(z) > \gamma_N \right\rbrace}_{\encircle{$*$}} \\
\end{align*}

We will bound $\mathbb{P}_Z\encircle{$*$} = 0$ using Chebyshev's inequality.
Noting that 
\begin{align*}
    \frac{\hat{q}_N(z;\XN)}{p(z)} 
    &= \frac{\hat{q}_N(z;{\XN}')}{p(z)} - \frac{1}{N}\frac{q(z|X'_1)}{p(z)} + \frac{1}{N}\frac{q(z|X_1)}{p(z)}, \\
\end{align*}
and using the fact that $\frac{q(z|X_1)}{p(z)} > 0$ it follows that
\begin{align*}
    \encircle{$*$} \implies &\gamma_N - \frac{1}{N}\frac{q(z|X'_1)}{p(z)} + \frac{1}{N}\frac{q(z|X_1)}{p(z)} < \delta_N \\
    \iff & (\gamma_N - \delta_N)N + \frac{q(z|X_1)}{p(z)} < \frac{q(z|X'_1)}{p(z)}\\
    \implies & (\gamma_N-\delta_N)N < \frac{q(z|X'_1)}{p(z)}  \\
\implies & (\gamma_N-\delta_N)N - 1 < \frac{q(z|X'_1)}{p(z)} - 1.
\end{align*}
where the penultimate line follows from the fact that $q(z|X_1)/p(z)\geq0$. It follows that
\begin{align*}
    \mathbb{P}_Z\encircle{$*$} &\leq \mathbb{P}_Z\left\lbrace \frac{q(z|X'_1)}{p(z)} - 1 > (\gamma_N-\delta_N)N - 1 \right\rbrace \\
    &\leq \mathbb{P}_Z\left\lbrace \left|\frac{q(z|X'_1)}{p(z)} - 1\right| > (\gamma_N-\delta_N)N - 1 \right\rbrace. \\
\end{align*}

Denote by $\sigma^2(X) = \V_Z\left[\frac{q(z|X)}{p(z)}\right] = \E_Z \frac{q^2(z|X)}{p^2(z)} - 1 \leq C$.
We have by Chebyshev that for any $t>0$,
\begin{align*}
    &\mathbb{P}_Z\left\lbrace \left|\frac{q(z|X)}{p(z)} - 1\right| > t \right\rbrace \leq \frac{\sigma^2(X)}{t^2} \\
\end{align*}
and so setting $t=(\gamma_N-\delta_N)N - 1$ yields
\begin{align*}
    \mathbb{P}_Z\encircle{$*$} &\leq \frac{\sigma^2(X)}{\left((\gamma_N-\delta_N)N - 1\right)^2} \leq \frac{C}{\left((\gamma_N-\delta_N)N - 1\right)^2}
\end{align*}

It follow that
\begin{align*}
    \encircle{B} &\leq f_0(0) \frac{C}{\left((\gamma_N-\delta_N)N - 1\right)^2}
\end{align*}

\encircle{C}: Similar to the proof of Theorem \ref{thm:convergence-rate-general}, we can upper bound this term by the product of two terms, one of which is independent of the choice of divergence.
The other term will be treated separately for each divergence considered.

\begin{align*}
    \encircle{C} &= \int {f_0}\left( \frac{\hat{q}_N(z;\XN)}{p(z)}\right) - {f_0}\left( \frac{\hat{q}_N(z;{\XN}')}{p(z)} \right) \mathds{1}_{\left\lbrace \frac{\hat{q}_N(z;\XN)}{p(z)} > \delta_N \right\rbrace} dP_Z(z) \\
    &\leq \int \left(\frac{\hat{q}_N(z;\XN)}{p(z)} - \frac{\hat{q}_N(z;{\XN}')}{p(z)}\right) {f_0}'\left(\frac{\hat{q}_N(z;\XN)}{p(z)} \right) \mathds{1}_{\left\lbrace \frac{\hat{q}_N(z;\XN)}{p(z)} > \delta_N \right\rbrace} dP_Z(z) 
    \quad \text{(Convexity of ${f_0}$)}
    \\
    &= \int \frac{1}{N}\frac{q(z|X_1) - q(z|X'_1)}{p(z)}  {f_0}'\left(\frac{\hat{q}_N(z;\XN)}{p(z)} \right) \mathds{1}_{\left\lbrace \frac{\hat{q}_N(z;\XN)}{p(z)} > \delta_N \right\rbrace} dP_Z(z) \\
    &\leq \left\| \frac{1}{N}\frac{q(z|X_1) - q(z|X'_1)}{p(z)}\right\|_{L_2(P_Z)}
    \left\| {f_0}'\left(\frac{\hat{q}_N(z;\XN)}{p(z)} \right) \mathds{1}_{\left\lbrace \frac{\hat{q}_N(z;\XN)}{p(z)} > \delta_N \right\rbrace} \right\|_{L_2(P_Z)} 
    \quad \text{(Cauchy-Schwartz)}
    \\
    &\leq \frac{2(C+1)}{N} \underbrace{\sqrt{\int {f_0}'^2\left(\frac{\hat{q}_N(z;\XN)}{p(z)} \right) \mathds{1}_{\left\lbrace \frac{\hat{q}_N(z;\XN)}{p(z)} > \delta_N \right\rbrace} p(z) dz }}_{\encircle{$*$}}
    \qquad \text{(Boundedness of ${\scriptscriptstyle\left\| \frac{q(z|x)}{p(z)}\right\|_{L_2(P_Z)}}$)}
    \\
\end{align*}
The term \encircle{$*$} will be treated separately for each divergence.

\paragraph{$\KL$:}

By Lemma \ref{lemma:concave-upper-bound-kl}, there exists a function $h_{\delta_N}(x)$ that is positive and concave on $[0, \infty)$ and is an upper bound of $f_0'^2(x)$ on $[\delta_N, \infty)$ with $h_{\delta_N}(1) = \log^2(\delta_N) + \frac{2}{e}$.

\begin{align*}
    \encircle{$*$}^2
    &\leq \int h_{\delta_N}\left(\frac{\hat{q}_N(z;\XN)}{p(z)} \right) \mathds{1}_{\left\lbrace \frac{\hat{q}_N(z;\XN)}{p(z)} > \delta_N \right\rbrace} p(z) dz
    && \text{($h_{\delta_N}$ upper bounds $f'^2$ on $(\delta_N, \infty)$)}
    \\
    &\leq \int h_{\delta_N}\left(\frac{\hat{q}_N(z;\XN)}{p(z)} \right) p(z) dz
    && \text{($h_{\delta_N}$ non-negative on $[0, \infty)$)}
    \\
    &\leq  h_{\delta_N}\left( \int \frac{\hat{q}_N(z;\XN)}{p(z)} p(z) dz \right) 
    && \text{($h_{\delta_N}$ concave)}
    \\
    &= h_{\delta_N}\left( 1\right) \\
    &= \log^2(\delta_N) + \frac{2}{e}\\
    \implies \encircle{$C$} &\leq \frac{2(C+1)}{N}\sqrt{\log^2(\delta_N) + \frac{2}{e}}.
\end{align*}

Putting together the separate integrals and setting $\delta_N = \frac{1}{N^{2/3}}$ and $\gamma_N = \frac{2}{N^{2/3}}$ , we have that

\begin{align*}
    &D_f\left( \hat{Q}_Z^N(\XN) \| P_Z\right) - D_f\left( \hat{Q}_Z^N({\XN}') \| P_Z\right) \\
    &= \encircle{A} + \encircle{B} + \encircle{C} \\
    &\leq f_0(0) - f_0\left(\gamma_N\right) +  \frac{f_0(0)C}{\left((\gamma_N-\delta_N)N - 1\right)^2} + \frac{2(C+1)}{N} \sqrt{\log^2(\delta_N) + \frac{2}{e} } \\
    &= \gamma_N - \gamma_N \log \gamma_N + \frac{f_0(0)C}{\left((\gamma_N-\delta_N)N - 1\right)^2} + \frac{2(C+1)}{N}  \sqrt{\log^2(\delta_N) + \frac{2}{e}}\\
    &=\frac{2}{N^{2/3}} - \frac{2}{N^{2/3}} \log\left(\frac{2}{N^{2/3}} \right) + \frac{f_0(0)C}{\left(N^{1/3} - 1\right)^2} + \frac{2(C+1)}{N} \sqrt{\frac{4}{9}\log^2(N) + \frac{2}{e}} 
    \\
    &\leq\frac{2}{N^{2/3}} - \frac{2}{N^{2/3}} \log\left(\frac{2}{N^{2/3}} \right) + \frac{9f_0(0)C}{4N^{2/3}} + \frac{2(C+1)}{N} \sqrt{\frac{4}{9}\log^2(N) + \frac{2}{e}} 
    \\
    &= \frac{k_1}{N^{2/3}} + \frac{k_2 \log N}{N^{2/3}} + \frac{k_3 \sqrt{\log^2N + \frac{9}{2e}}}{N}\\
    &\leq (k_1+k_2+2k_3)\frac{\log N}{N^{2/3}}
\end{align*}
where $k_1, k_2$ and $k_3$ are constants depending on $C$.
The second inequality holds if $N^{1/3}-1 > \frac{N^{1/3}}{3} \iff N>\left(\frac{3}{2}\right)^3 < 4$ and the third inequality holds if $N\geq 4$

The assumption that $\delta_N, \gamma_N \leq 1$ holds if $N>2^{3/2}$ and so holds if $N\geq3$.

This leads to $Nc_N^2 = \frac{\log^2N}{N^{1/3}}$ for $N> 3$.

\paragraph{Squared Hellinger.}

In this case similar reasoning to the other divergences leads to a bound that is worse than $O\left(\frac{1}{\sqrt{N}}\right)$ and thus $Nc^2_N$ is bigger than $O(1)$ leading to a trivial concentration result.

\paragraph{$\alpha$-divergence with $\alpha\in(\frac{1}{3},1)$.}

Following similar reasoning to the proof of Theorem \ref{thm:convergence-rate-general} for the $\alpha$-divergence case, we use the function $h_{\delta_N}(x)$ provided by Lemma \ref{lemma:upper-bound-alpha} to derive the following upper bound:

\begin{align*}
    \encircle{C} &\leq \frac{2(C+1)}{N} \cdot \frac{2\sqrt{1+(C+1)^2}\left(\delta_N^\frac{\alpha-1}{2} - 1\right)}{(\alpha - 1) (\delta_N- 1)}.
\end{align*}

Setting $\delta_N = \frac{1}{N^\frac{4}{\alpha+5}}$ and $\gamma_N = \frac{2}{N^\frac{4}{\alpha+5}}$,

\begin{align*}
    &D_f\left( \hat{Q}_Z^N(\XN) \| P_Z\right) - D_f\left( \hat{Q}_Z^N({\XN}') \| P_Z\right) \\
    &= \encircle{A} + \encircle{B} + \encircle{C} \\
    &\leq f_0(0) - f_0\left(\gamma_N\right) + \frac{f_0(0)C}{\left((\gamma_N-\delta_N)N - 1\right)^2} + \frac{2(C+1)}{N}\frac{2\sqrt{1+(C+1)^2}\left(\delta_N^\frac{\alpha-1}{2} - 1\right)}{(1-\alpha) (1-\delta_N)} \\
    &\leq f_0(0) - f_0\left(\gamma_N\right) + \frac{t^2f_0(0)C}{(t-1)^2(\gamma_N-\delta_N)^2N^2} + \frac{2(C+1)}{N}\frac{2\sqrt{1+(C+1)^2}\left(\delta_N^\frac{\alpha-1}{2} - 1\right)}{(1-\alpha) (1-\delta_N)} \\
    &\leq f_0(0) - f_0\left(\gamma_N\right) + \frac{t^2f_0(0)C}{(t-1)^2(\gamma_N-\delta_N)^2N^2} + \frac{2(C+1)}{N}\frac{4\sqrt{1+(C+1)^2}\delta_N^\frac{\alpha-1}{2}}{(1-\alpha)} \\
    &\leq k_1 \gamma_N^{\frac{\alpha+1}{2}} + k_2 \gamma_N +\frac{k_3}{(\gamma_N-\delta_N)^2N^2} + \frac{k_4 \delta_N^\frac{\alpha-1}{2}}{N} \\
    &= \frac{k_1}{N^{\frac{2\alpha+2}{\alpha+5}}} + \frac{k_2}{N^\frac{4}{\alpha+5}} + \frac{k_3}{N^{\frac{2\alpha -2}{\alpha+5}}} +\frac{k_4}{N^{\frac{3\alpha+3}{\alpha+5}}} \\
    &\leq \frac{k_1+k_2+k_3 + k_4}{N^{\frac{2\alpha+2}{\alpha+5}}}
\end{align*}
where $t$ is any positive number and where the second inequality holds if $N^\frac{2\alpha+2}{\alpha+5} - 1 > \frac{N^\frac{2\alpha+2}{\alpha+5}}{t} \iff N > (\frac{t}{t-1})^{\frac{\alpha+5}{2\alpha+21}}$.
For $\alpha \in (\frac{1}{3}, 1)$ we have $\frac{\alpha+5}{2\alpha+2} \in (\frac{3}{2}, 2)$. 
If we take $t=100$ then $N> 1$ suffices for any $\alpha$.

The third inequality holds if $1-\delta_N > \frac{1}{2} \iff N>2^\frac{\alpha+5}{4}$ and so holds if $N>3$.

The assumption that $\delta_N, \gamma_N \leq 1$ holds if $N>4^\frac{\alpha+5}{4}\leq8$ and so holds if $N>8$.

Thus, this leads to $Nc_N^2 = \frac{k}{N^{\frac{3\alpha - 1}{\alpha+5}}}$ for $N>8$.

\paragraph{Jensen-Shannon.}

Following similar reasoning to the proof of Theorem \ref{thm:convergence-rate-general} for the $\alpha$-divergence case, we use the function $h_{\delta_N}(x)$ provided by Lemma \ref{lemma:upper-bound-JS} to derive the following upper bound:

\begin{align*}
    \encircle{C} &\leq \frac{2(C+1)}{N} \cdot \sqrt{5}\log\left(\frac{1}{\delta_N}\right).
\end{align*}

Setting $\delta_N = \frac{1}{N^{2/3}}$ and $\gamma_N = \frac{2}{N^{2/3}}$,

\begin{align*}
    &D_f\left( \hat{Q}_Z^N(\XN) \| P_Z\right) - D_f\left( \hat{Q}_Z^N({\XN}') \| P_Z\right) \\
    &= \encircle{A} + \encircle{B} + \encircle{C} \\
    &\leq f_0(0) - f_0\left(\gamma_N\right) + \frac{f_0(0)C}{\left((\gamma_N-\delta_N)N - 1\right)^2} + \frac{2(C+1)}{N} \cdot \log\left(\frac{1}{\delta_N}\right) \\
    &\leq \gamma_N \log\left(\frac{1+\gamma_N}{2\gamma_N}\right) + \log(1+\gamma_N) + \frac{f_0(0)C}{\left((\gamma_N-\delta_N)N - 1\right)^2} + \frac{2(C+1)}{N} \cdot \log\left(\frac{1}{\delta_N}\right).\\
\end{align*}

Using the fact that $\log(1+\gamma_N)\leq \gamma_N$, we obtain the following upper bound:

\begin{align*}
    &\leq \gamma_N^2 + \gamma_N(1-\log 2 ) - \gamma_N \log \gamma_N + \frac{f_0(0)C}{\left((\gamma_N-\delta_N)N - 1\right)^2} + \frac{2(C+1)}{N} \cdot \log\left(\frac{1}{\delta_N}\right)\\
    &= \frac{k_1}{N^{4/3}} + \frac{k_2}{N^{2/3}} + \frac{k_3\log N}{N^{2/3}} + \frac{k_4}{(N^{1/3} - 1)^2} +\frac{k_5 \log N }{ N^{2/3}} \\
    &= \frac{k_1}{N^{4/3}} + \frac{k_2}{N^{2/3}} + \frac{k_3\log N}{N^{2/3}} + \frac{k_4}{(N^{1/3} - 1)^2} +\frac{k_5 \log N }{ N^{2/3}} \\
    &\leq \frac{k_1}{N^{4/3}} + \frac{k_2}{N^{2/3}} + \frac{k_3\log N}{N^{2/3}} + \frac{100k_4}{81N^{2/3}} +\frac{k_5 \log N }{ N^{2/3}} \\
    &\leq (k_1+k_2+k_3+k'_4 + k_5)\frac{\log N}{N^{2/3}}
\end{align*}
where the penultimate inequality holds if $N^{1/3}-1 > \frac{N^{1/3}}{10} \iff N>\left(\frac{10}{9}\right)^3$ which is satisfied if $N>1$ and the last inequality is true if $N>1$.

The assumption that $\delta_N, \gamma_N \leq 1$ holds if $N>2^{3/2}$ and so holds if $N\geq3$.

This leads to $Nc_N^2 = \frac{\log^2N}{N^{1/3}}$ for $N>2$.

\paragraph{$f_\beta$-divergence, $\beta\in(\frac{1}{2},1)$.}

Following similar reasoning to the proof of Theorem \ref{thm:convergence-rate-general} for the $\alpha$-divergence case, we use the function $h_{\delta_N}(x)$ provided by Lemma \ref{lemma:upper-bound-f-beta} to derive the following upper bound:

\begin{align*}
    \encircle{C} &\leq \frac{2(C+1)}{N} \cdot \frac{\beta}{1-\beta}\cdot 2^{\frac{2+\beta}{2\beta}}\delta_N^{\beta-1}.
\end{align*}

Setting $\delta_N = \frac{1}{N^{2/3}}$ and $\gamma_N =\frac{2}{N^{2/3}}$,

\begin{align*}
    &D_f\left( \hat{Q}_Z^N(\XN) \| P_Z\right) - D_f\left( \hat{Q}_Z^N({\XN}') \| P_Z\right) \\
    &= \encircle{A} + \encircle{B} + \encircle{C} \\
    &\leq f_0(0) - f_0\left(\gamma_N\right) + \frac{f_0(0)C}{\left((\gamma_N-\delta_N)N - 1\right)^2} + \frac{\beta}{1-\beta}\cdot 2^{\frac{2+\beta}{2\beta}}\delta_N^{\beta-1} \\
    &\leq \frac{\beta}{\beta-1}2^{\frac{1-\beta}{\beta}}\gamma_N + \frac{f_0(0)C}{\left((\gamma_N-\delta_N)N - 1\right)^2}+ \frac{\beta}{1-\beta}\cdot 2^{\frac{2+\beta}{2\beta}} \frac{ \delta^{\beta-1}}{N} \\
    &= \frac{k_1}{N^{2/3}} + \frac{k_2}{(N^{1/3} - 1)^2} + \frac{k_3}{N^\frac{2\beta + 1}{3}}\\
    &\leq \frac{k_1}{N^{2/3}} + \frac{100k_2}{81N^{2/3}} + \frac{k_3}{N^\frac{2\beta + 1}{3}}\\
    &\leq \frac{k_1+k'_2 + k_3}{N^{2/3}}
\end{align*}
where the penultimate inequality holds if $N^{1/3}-1 > \frac{N^{1/3}}{10} \iff N>\left(\frac{10}{9}\right)^3$ which is satisfied if $N>1$.

The assumption that $\delta_N, \gamma_N \leq 1$ holds if $N>2^{3/2}$ and so holds if $N\geq3$.

This leads to $Nc_N^2 = \frac{1}{N^{1/3}}$ for $N>2$.

\end{proof}
\subsection{Full statement and proof of Theorem \ref{thm:mc-variance}}\label{appendix:full-statment-proof-mc}

The statement of Theorem~\ref{thm:mc-variance} in the main text was simplified for brevity. 
Below is the full statement, followed by its proof.

\setcounter{theorem}{3}

\begin{theorem}
For any $\pi$,
\begin{align*}
    \E_{\ZM, \XN} \bigl[\hat{D}^M_f( \hat{Q}_Z^N \| P_Z)\bigr]
    &=\E_{\XN} \left[ D_f\left( \hat{Q}^N_Z \| P_Z \right)\right].
\end{align*}
If either of the following conditions are satisfied:
\begin{align*}
&(i) \ \pi(z|\XN) = p(z),& 
&\textstyle{\E_X \left\| f\left( \frac{q(z|X)}{p(z)}\right) \right\|^2_{L_2(P_Z)}  < \infty,}&
&\textstyle{\E_X \left\| \frac{q(z|X)}{p(z)} \right\|^2_{L_2(P_Z)} < \infty} \\
&(ii) \  \pi(z | \XN) = \hat{q}_N(z),&
&\textstyle{\E_X \left\| f\left( \frac{q(z|X)}{p(z)}\right)\frac{p(z)}{q(z|X)}\right\|^2_{L_2(Q_{Z|X})} < \infty,}&
&\textstyle{\E_X \left\| \frac{p(z)}{q(z|X)} \right\|^2_{L_2(Q_{Z|X})} < \infty}
\end{align*}
then, denoting by $\psi(N)$ the rate given in Table~\ref{table:concentration}, we have
\begin{align*}
    \text{Var}_{\ZM, \XN} \left[\hat{D}^M_f( \hat{Q}_Z^N \| P_Z)  \right] = 
    O\left(M^{-1}\right) + O\left( \psi(N)^2 \right) 
\end{align*}
\end{theorem}

In proving Theorem~\ref{thm:mc-variance} we will make use of the following lemma.

\begin{lemma}\label{lemma:f0-x-convex}
For any $f_0(x)$,
the functions $f_0(x)^2$ and $\frac{f_0(x)^2}{x}$ are convex on $(0, \infty)$.
\end{lemma}
\begin{proof}
To see that $f_0(x)^2$ is convex, observe that 
\begin{align*}
    \frac{d^2}{dx^2} f_0(x)^2 = 2\left( f_0(x) f_0''(x) + f_0'(x)^2 \right)
\end{align*}
All of these terms are postive for $x > 0$.
Indeed, since $f_0(x)$ is convex for $x > 0$, $f_0''(x) \geq 0$.  
By construction of $f_0$, $f_0(x) \geq 0$ for $x > 0$. 
Thus $f_0(x)^2$ has non-negative derivative and is thus convex on $(0, \infty)$.

To see that $\frac{f_0(x)^2}{x}$ is convex, observe that

\begin{align*}
    \frac{d^2}{dx^2} \frac{f_0(x)^2}{x} = \frac{2}{x}\left( f_0(x) f_0''(x) + \left(f_0'(x) - \frac{f_0(x)}{x}\right)^2 \right).
\end{align*}
By the same arguments above, this is positive for $x > 0$ and thus $\frac{f_0(x)^2}{x}$ is convex for $x>0$.
\end{proof}

\begin{proof}(Theorem \ref{thm:mc-variance})
For the expectation, observe that

\begin{align*}
    \E_{\ZM, \XN} \hat{D}^M_f( \hat{Q}_Z^N \| P_Z)
    &=\E_{\XN} \left[ \E_{\ZM \overset{\text{\emph{i.i.d.}}}{\sim} \pi(z | \XN)} \hat{D}^M_f( \hat{Q}_Z^N \| P_Z)\right] \\
    &= \E_{\XN} \left[ \E_{z \sim \pi(z | \XN)} f\left(\frac{\hat{q}_N(z)}{p(z)} \right) \frac{p(z)}{\pi(z|\XN)} \right] \\
    &=\E_{\XN} \left[ D_f\left( \hat{Q}^N_Z \| P_Z \right)\right].
\end{align*}

For the variance, by the law of total variance we have that

\begin{align*}
    &\text{Var}_{\ZM, \XN} \left[\hat{D}^M_f( \hat{Q}_Z^N \| P_Z)  \right]  \\
    &= \E_{\XN} \text{Var}_{\ZM \overset{\text{\emph{i.i.d.}}}{\sim} \pi(z|\XN)} \hat{D}^M_f( \hat{Q}_Z^N \| P_Z) + \text{Var}_{\XN} \mathbb{E}_{\ZM \overset{\text{\emph{i.i.d.}}}{\sim} \pi(z|\XN)} \hat{D}^M_f( \hat{Q}_Z^N \| P_Z)
    \\
    &=\frac{1}{M} \underbrace{\E_{\XN} \text{Var}_{\pi(z|\XN)} \left[ f\left( \frac{\hat{q}_N(z)}{p(z)} \right) \frac{p(z)}{\pi\left(z|\XN\right)} \right]}_{(i)}  + \underbrace{\text{Var}_{\XN} \left[D_f\left( \hat{Q}^N_Z \| P_Z \right)\right]}_{(ii)}.
\end{align*}

Consider term $(ii)$.
The concentration results of Theorem \ref{thm:concentration} imply bounds on $(ii)$, since for a random variable $X$,
\begin{align*}
    \text{Var}X &= \mathbb{E} (X - EX)^2 \\
    &= \int_0^\infty \mathbb{P}\left( (X - \mathbb{E} X)^2 > t \right) dt \\
    &= \int_0^\infty \mathbb{P} \left( \left| X - \mathbb{E} X \right| > \sqrt{t} \right) dt.
\end{align*}
It follows therefore that
\begin{align*}
    \text{Var}_{\XN} \left[D_f\left( \hat{Q}^N_Z \| P_Z \right)\right] 
    &\leq \int_0^\infty 2 \exp\left( -\frac{k}{\psi(N)^2}t \right) dt\\
    &= O\left(\psi(N)^2 \right)
\end{align*}
where $\psi(N)$ is given by Table~\ref{table:concentration}.

Next we consider $(i)$ and show that it is bounded independent of $N$, and so the component of the variance due to this term is $O\left(\frac{1}{M}\right)$.
In the case that $\pi(z|\XN) = p(z)$,
\begin{align*}
(i) &\leq \E_{\XN} \E_{p(z)} \left[ f\left( \frac{\hat{q}_N(z)}{p(z)} \right)^2\right]\\
&= \E_{\XN} \E_{p(z)} \left[ \left( f_0\left( \frac{\hat{q}_N(z)}{p(z)} \right) + f'(1) \left( \frac{\hat{q}_N(z)}{p(z)} - 1 \right) \right)^2\right] \\
&\leq \E_{\XN} \E_{p(z)} \left[  f_0\left( \frac{\hat{q}_N(z)}{p(z)} \right)^2\right] 
+
 f'(1)^2 \E_{\XN} \E_{p(z)} \left[  \left( \frac{\hat{q}_N(z)}{p(z)} - 1 \right)^2\right] \\
 & \qquad + 2f'(1) \sqrt{\E_{\XN} \E_{p(z)} \left[  f_0\left( \frac{\hat{q}_N(z)}{p(z)} \right)^2\right] } \times \sqrt{\E_{\XN} \E_{p(z)} \left[  \left( \frac{\hat{q}_N(z)}{p(z)} - 1 \right)^2\right]} \\
&\leq \E_{X} \E_{p(z)} \left[  f_0\left( \frac{q(z|X)}{p(z)} \right)^2\right] 
+
 f'(1)^2 \E_{X} \E_{p(z)} \left[  \left( \frac{q(z|X)}{p(z)} - 1 \right)^2\right] \\
 & \qquad + 2f'(1) \sqrt{\E_{X} \E_{p(z)} \left[  f_0\left( \frac{q(z|X)}{p(z)} \right)^2\right] } \times \sqrt{\E_{X} \E_{p(z)} \left[  \left( \frac{q(z|X)}{p(z)} - 1 \right)^2\right]} \\
\end{align*}
The penultimate inequality follows by application of Cauchy-Schwartz.
The last inequality follows by Proposition \ref{prop:upper-bound} applied to $D_{f_0^2}$ and $D_{(x-1)^2}$, 
using the fact that the functions $f_0^2(x)$ and $(x-1)^2$ are convex and are zero at $x=1$ (see Lemma \ref{lemma:f0-x-convex}).
By assumption, $\E_{X} \E_{p(z)} \left[  \left( \frac{q(z|X)}{p(z)} - 1 \right)^2\right] < \infty$. 
Consider the other term:
\begin{align*}
    \E_{X} \E_{p(z)} \left[  f_0\left( \frac{q(z|X)}{p(z)} \right)^2\right]
    &= \E_{X} \E_{p(z)} \left[ \left( f\left( \frac{q(z|X)}{p(z)}\right) - f'(1) \left(\frac{q(z|X)}{p(z)} - 1 \right) \right)^2\right] \\
    &\leq \E_{X} \E_{p(z)} \left[  f\left( \frac{q(z|X)}{p(z)} \right)^2\right] 
    +
    f'(1)^2 \E_{X} \E_{p(z)} \left[  \left( \frac{q(z|X)}{p(z)} - 1 \right)^2\right] \\
    & \qquad + 2f'(1) \sqrt{\E_{X} \E_{p(z)} \left[  f\left( \frac{q(z|X)}{p(z)} \right)^2\right] } \times \sqrt{\E_{X} \E_{p(z)} \left[  \left( \frac{q(z|X)}{p(z)} - 1 \right)^2\right]} \\
    & < \infty
\end{align*}
The inequality follows by Cauchy-Schwartz.
All terms are finite by assumption.
Thus $(i) \leq K < \infty$ for some $K$ independent of $N$.

Now consider the case that $\pi(z|\XN) = \hat{q}_N(z)$. 
Then, following similar (but algebraically more tedious) reasoning to the previous case, it can be shown that

\begin{align*}
(i)
&\leq \E_{X} \E_{p(z)} \left[  f_0\left( \frac{q(z|X)}{p(z)} \right)^2 \frac{p(z)}{q(z|X)}\right] 
+
 f'(1)^2 \E_{X} \E_{p(z)} \left[  \left( \sqrt{\frac{q(z|X)}{p(z)}} - \sqrt{\frac{p(z)}{q(z|X)}} \right)^2\right] \\
 & \qquad + 2f'(1) \sqrt{\E_{X} \E_{p(z)} \left[  f_0\left( \frac{q(z|X)}{p(z)} \right)^2 \frac{p(z)}{q(z|X)}\right] } \times \sqrt{\E_{X} \E_{p(z)} \left[  \left( \sqrt{\frac{q(z|X)}{p(z)}} - \sqrt{\frac{p(z)}{q(z|X)}} \right)^2\right]} \\
\end{align*}
where Proposition \ref{prop:upper-bound} is applied to $D_{\frac{f_0^2(x)}{x}}$ and $D_{(\sqrt{x}- \frac{1}{\sqrt{x}})^2}$, 
using the fact that the functions $f_0^2(x)/x$ and $(\sqrt{x}- \frac{1}{\sqrt{x}})^2$ are convex and are zero at $x=1$ (see Lemma \ref{lemma:f0-x-convex}).
Noting that
\begin{align*}
    \E_{X} \E_{p(z)} \left[  \left( \sqrt{\frac{q(z|X)}{p(z)}} - \sqrt{\frac{p(z)}{q(z|X)}} \right)^2\right]
    &=\E_{X} \E_{p(z)} \left[ \frac{q(z|X)}{p(z)} + \frac{p(z)}{q(z|X)} - 2 \right] \\
    &=\E_{X} \E_{p(z)} \left[ \frac{p(z)}{q(z|X)} - 1 \right] < \infty
\end{align*}
where the inequality holds by assumption, it follows that
\begin{align*}
    &\E_{X} \E_{p(z)} \left[  f_0\left( \frac{q(z|X)}{p(z)} \right)^2 \frac{p(z)}{q(z|X)}\right]\\
    &\leq \E_{X} \E_{p(z)} \left[  f\left( \frac{q(z|X)}{p(z)} \right)^2 \frac{p(z)}{q(z|X)}\right] 
    +
    f'(1)^2 \E_{X} \E_{p(z)} \left[  \left( \sqrt{\frac{q(z|X)}{p(z)}} - \sqrt{\frac{p(z)}{q(z|X)}} \right)^2\right] \\
    & \qquad + 2f'(1) \sqrt{\E_{X} \E_{p(z)} \left[  f\left( \frac{q(z|X)}{p(z)} \right)^2 \frac{p(z)}{q(z|X)}\right] } \times \sqrt{\E_{X} \E_{p(z)} \left[  \left( \sqrt{\frac{q(z|X)}{p(z)}} - \sqrt{\frac{p(z)}{q(z|X)}} \right)^2\right]} \\
    & < \infty.
\end{align*}
where the first inequality holds by the definition of $f_0$ and Cauchy-Schwartz. 

Thus $(i) \leq K < \infty$ for some $K$ independent of $N$ in both cases of $\pi$.
\end{proof}
\subsection{Elaboration of Section \ref{subsection:discussion-assumptions}: satisfaction of assumptions of theorems}\label{appendix:discussion-constraints}

Suppose that ${P_Z}$ is  ${\mathcal{N}\left(0, I_d\right)}$ and ${Q_{Z|X}}$ is  ${\mathcal{N}\left( \mu(X), \Sigma(X)\right)}$ with $\Sigma$ diagonal. 
Suppose further that there exist constants $K, \epsilon > 0$ such that ${\| \mu(X)\| \leq K}$ and ${\Sigma_{ii}(X) \in [\epsilon, 1]}$ for all $i$.

By Lemma~\ref{lemma:chi-squared-closed-form}, it holds that $\chi^2\bigl( Q_{Z|x}, P_Z\bigr) < \infty$ for all $x\in\mathcal{X}$. 
By compactness of the sets in which $\mu(X)$ and $\Sigma(X)$ take value, it follows that there exists $C<\infty$ such that $\chi^2\bigl( Q_{Z|x}, P_Z\bigr) \leq C$ and thus the setting of Theorem~\ref{thm:concentration} holds.

A similar argument based on compactness shows that the density ratio is uniformly bounded in $z$ and $x$: $q(z|x)/p(z) \leq C'$ for some $C'<\infty$. 
It therefore follows that the condition of Theorem~\ref{thm:convergence-rate-general} holds: $\int q^4(z|x)/p^4(z) dP(z) < {C'}^4 < \infty$.

We conjecture that the strong boundedness assumptions on $\mu(X)$ and $\Sigma(X)$ also imply the setting of Theorem~\ref{thm:fast-KL-rate} $\E_{X}\bigl[\chi^2\bigl(Q_{Z|X}, Q_Z\bigr)\bigr] < \infty$.
Since the divergence $Q_Z$ explicitly depends on the data distribution, this is more difficult to verify than the conditions of Theorems~\ref{thm:convergence-rate-general} and \ref{thm:concentration}.

The crude upper bound provided by convexity
\[
\E_{X}\bigl[\chi^2\bigl(Q_{Z|X}, Q_Z\bigr)\bigr] \leq  \E_{X}\E_{X'}\bigl[\chi^2\bigl(Q_{Z|X}, Q_{Z|X'}\bigr)\bigr]
\]
provides a sufficient (but very strong) set of assumptions under which it holds. 
Finiteness of the right hand side above would be implied, for instance, by demanding that ${\| \mu(X)\| \leq K}$ and ${\Sigma_{ii}(X) \in [\frac{1}{2}+\epsilon, 1]}$ for all $i$.
\section{Empirical evaluation: further details}\label{appendix:empirical-evaluation-details}

In this section with give further details about the synthetic and real-data experiments presented in Section \ref{sec:experiments}.

\subsection{Synthetic experiments}

\subsubsection{Analytical expressions for divergences between two Gaussians}\label{appendix:toy-exps}

The closed form expression for the $\chi^2$-divergence between two $d$-variate normal distributions can be found in Lemma 1 of~\cite{NielsenN14}:
\begin{lemma}\label{lemma:chi-squared-closed-form}
\begin{align*}
&\chi^2\bigl( \mathcal{N}(\mu_1, \Sigma_1), 
\mathcal{N}(\mu_2, \Sigma_2)\bigr)
=
\frac{\mathrm{det}(\Sigma_1^{-1})}{\sqrt{\mathrm{det}(2\Sigma_1^{-1} - \Sigma_2^{-1})\mathrm{det}(\Sigma_2^{-1})}}
\exp\left(
\frac12\mu_2'\Sigma_2^{-1}\mu_2 
-\mu_1'\Sigma_1^{-1}\mu_1 
\right)
\times\\
&\times\exp\left(
-\frac14(2\mu_1' \Sigma_1^{-1} - \mu_2' \Sigma_2^{-1})
\bigl(\frac12 \Sigma_2^{-1} - \Sigma_1^{-1}\bigr)^{-1}
(2\Sigma_1^{-1}\mu_1 - \Sigma_2^{-1}\mu_2)
\right) - 1.
\end{align*}
\end{lemma}
As a corollary, the following also holds:
\begin{corollary}
Chi square divergence between two $d$-variate Gaussian distributions both having covariance matrices proportional to identity can be computed as:
\[
\chi^2\bigl( \mathcal{N}(\mu, \sigma^2 I_d), \mathcal{N}(0, \beta^2 I_d)\bigr)
=
\left(\frac{\beta^2}{\sigma^2\sqrt{2\beta^2/\sigma^2 - 1}}\right)^d
e^{\frac{\|\mu\|^2}{2\beta^2 - \sigma^2}}
- 1
\]
assuming $2\beta^{2} > \sigma^{2}$. 
Otherwise the divergence is infinite.
\end{corollary}

The squared Hellinger divergence between two Gaussians is given in \cite{pardo2005statistical}:

\begin{lemma}
\begin{align*}
&H^2\bigl( \mathcal{N}(\mu_1, \Sigma_1), 
\mathcal{N}(\mu_2, \Sigma_2)\bigr) \\
&\qquad = 1 - \frac{ \det (\Sigma_1)^{1/4} \det (\Sigma_2) ^{1/4}} { \det \left( \frac{\Sigma_1 + \Sigma_2}{2}\right)^{1/2} }
              \exp\left\{-\frac{1}{8}(\mu_1 - \mu_2)^T 
              \left(\frac{\Sigma_1 + \Sigma_2}{2}\right)^{-1}
              (\mu_1 - \mu_2)              
              \right\}.
\end{align*}
\end{lemma}

The $\KL$-divergence between two $d$-variate Gaussians is:
 
\begin{lemma}
\begin{align*}
\KL\bigl( \mathcal{N}(\mu_1, \Sigma_1), 
\mathcal{N}(\mu_2, \Sigma_2)\bigr) = \frac{1}{2} \left( \text{tr}\left(\Sigma_2^{-1} \Sigma_1\right)
+ (\mu_2  - \mu_1)^\intercal \Sigma_2^{-1}(\mu_2 - \mu_1) - d + \log\frac{|\Sigma_2|}{|\Sigma_1|}
\right).
\end{align*}
\end{lemma}

\subsubsection{Further experimental details}

We take $Q^\lambda_{Z|X=x} = \mathcal{N}\left(A_\lambda x + b_\lambda, \epsilon^2 I_d \right)$ and $P_X = \mathcal{N}\left(0, I_{20} \right)$.
This results in $Q^\lambda_Z = \mathcal{N}\left(b_\lambda,  A_\lambda A_\lambda^\intercal + \epsilon^2 I_d \right)$. We chose $\epsilon=0.5$ and used $\lambda \in [-2,2]$. $P_Z = \mathcal{N}(0, I_d)$.

$A_\lambda$ and $b_\lambda$ were determined as follows:
Define $A_1$ to be the $(d, 20)$-dimensional matrix with 1's on the main diagonal, and let $A_0$ be similarly sized matrix with entries randomly sampled i.i.d.\: unit Gaussians which is then normalised to have unit Frobenius norm. 
Let $v$ be a vector randomly sampled from the $d$-dimensional unit sphere.
We then set
$A_\lambda=\frac{1}{2} A_1 + \lambda A_0$ and $b_\lambda= \lambda v$.

$A_0$ and $v$ are sampled once for each dimension $d{\in}\{1,4,16\}$, such that the within each column of Figure~\ref{fig:synthetic-exps}, the distributions used are the same.

\subsection{Real-data experiments}\label{appendix:real-data-experiments-additional}

\subsubsection{Variational Autoencoders (VAEs) and Wasserstein Autoencoders (WAEs)}\label{appendix:intro-vae-wae}

Autoencoders are a general class of models typically used to learn compressed representations of high-dimensional data.
Given a \emph{data-space} $\mathcal{X}$ and low-dimensional \emph{latent space} $\mathcal{Z}$, the goal is to learn an \emph{encoder} mapping $\mathcal{X}\to\mathcal{Z}$ and \emph{generator} (or \emph{decoder}\footnote{In the VAE literature, the encoder and generator are sometimes referred to as the \emph{inference network} and \emph{likelihood model} respectively.}) mapping $\mathcal{Z}\to\mathcal{X}$.
The objectives used to train these two components always involve some kind of reconstruction loss measuring how corrupted a datum becomes after mapping through both the encoder and generator, and often some kind of regularization.

Representing by $\theta$ and $\eta$ the parameters of the encoder and generator respectively, the objective functions of VAEs and WAEs are:

\begin{align*}
    L^{\text{VAE}}(\theta, \eta) &= \E_X \left[ \E_{q_\theta(Z|X)} \log p_\eta(X|Z) + \KL\left( Q^\theta_{Z|X} \| P_Z) \right) \right]\\
    L^{\text{WAE}}(\theta, \eta) &= \E_X \E_{q_\theta(Z|X)} c(X, G_\eta(Z)) + \lambda \cdot D(Q^\theta_Z \| P_Z)
\end{align*}

For VAEs, both encoder $Q^\theta_{Z|X}$ and generator $p_\eta$ are \emph{stochastic} mappings taking an input and mapping it to a distribution over the output space.
In WAEs, only the encoder $Q^\theta_{Z|X}$ is stochastic, while the generator $G_\eta$ is deterministic.
$c$ is a cost function, $\lambda$ is a hyperparameter and $D$ is any divergence.

A common assumption made for VAEs is that the generator outputs a Gaussian distribution with fixed diagonal covariance and mean $\mu(z)$ that is a function of the input $z$.
In this case, the $\log p_\eta(X|z)$ term can be written as the $l^2_2$ (i.e. square of the $l_2$ distance) between $X$ and its reconstruction after encoding and re-generating $\mu(z)$.
If the cost function of the WAE is chosen to be $l^2_2$, then the left hand terms of the VAE and WAE losses are the same. 
That is, in this particular case, $L^{\text{VAE}}$ and $L^{\text{WAE}}$ differ only in their regularizers.

The penalty of the VAE was shown by \cite{hoffman2016elbo} to be equivalent to $\smash{\KL(Q^\theta_Z \| P_Z) + I(X,Z)}$ where $I(X,Z)$ is the mutual information of a sample and its encoding.
For the WAE penalty, there is a choice of which $\smash{D(Q^\theta_Z \| P_Z)}$ to use; it must only be possible to practically estimate it.
In the experiments used in this paper, we considered models trained with the Maximum Mean Discrepency (MMD) \cite{gretton2012kernel}, a kernel-based distance on distributions, and a divergence estimated using a GAN-style classifier \cite{goodfellow2014generative} leading to WAE-MMD and WAE-GAN respectively, following \cite{tolstikhin2017wasserstein}.

\subsubsection{Further experimental details}

We took a corpus of VAE, WAE-GAN and WAE-MMD models that had been trained with a large variety of hyperparameters including learning rate, latent dimension (32, 64, 128), architecture (ResNet/DCGAN), scalar factor for regulariser, and additional algorithm-specific hyperparameters: kernel bandwidth for WAE-MMD and learning rate of discriminator for WAE-GAN.
In total, 60 models were trained of each type (WAE-MMD, WAE-GAN and VAE) leading to 180 models in total.

The small subset of six models exposed in Figures~\ref{fig:real-exps} and \ref{fig:real-exps-hsq} were selected by a heuristic that we next describe. However, we note that qualitatively similar behaviour was found in all other models tested, and so the choice of models to display was somewhat arbitrary; we describe it nonetheless for completeness.

Recall that the objective functions of WAEs and VAEs both include a divergence between $Q^\theta_Z$ and $P_Z$.
We were interested in considering models from the two extremes of the distribution matching: some models in which $Q^\theta_Z$ and $P_Z$ were close, some in which they were distant.

To determine whether $Q^\theta_Z$ and $P_Z$ in a model are close, we made use of FID \cite{heusel2017gans} scores as a proxy that is independent of the particular divergences for training.
The FID score between two distributions over images is obtained by pushing both distributions through to an intermediate feature layer of the \emph{Inception} network.
The resulting push-through distributions are approximated with Gaussians and the \emph{Fr\'echet} distance between them is calculated.
Denote by $G_\#(Q^\theta_Z)$ the distribution over reconstructed images, $G_\#(P_Z)$ the distribution over model samples and $Q_X$ the data distribution, where $G$ is the generator and $\#$ denotes the push-through operator. 
The quantity $\text{FID}\left(Q_X, G_\#(Q^\theta_Z)\right)$ is a measure of quality (lower is better) of the reconstructed data,
while $\text{FID}\left(Q_X, G_\#(P_Z)\right)$ is a measure of quality of model samples.

The two FID scores being very different is an indication that $P_Z$ and $Q^\theta_Z$ are different.
In contrast, if the two FID scores are similar, we cannot conclude that $P_Z$ and $Q^\theta_Z$ are the same, though it provides some evidence towards that fact.
Therefore, in order to select a model in which matching between $P_Z$ and $Q^\theta_Z$ is poor, we pick one for which $\text{FID}\left(Q_X, G_\#(Q^\theta_Z)\right)$ is small but $\text{FID}\left(Q_X, G_\#(P_Z)\right)$ is large (good reconstructions; poor samples).
In order to select a model in which matching between $P_Z$ and $Q^\theta_Z$ is good, we pick one for both FIDs are small (good reconstructions; good samples). 
We will refer to these settings as \emph{poor matching} and \emph{good matching} respectively.

Our goal was to pick models according to the following criteria. 
The six chosen should include: two from each model class (VAE, WAE-GAN, WAE-MMD), of which one from each should exhibit poor matching and one good matching; two from each dimension $d\in\{32, 64, 128\}$; three with the ResNet architecture and three with the DCGAN architecture.
A set of models satisfying these criteria were selected by hand, but as noted previously we saw qualitatively similar results with the other models.

\subsubsection{Additional results for squared Hellinger distance}\label{appendix:sq-hellinger-results}

\begin{figure}
\begin{center}
\includegraphics[width=1.\textwidth, height=0.615\textwidth]{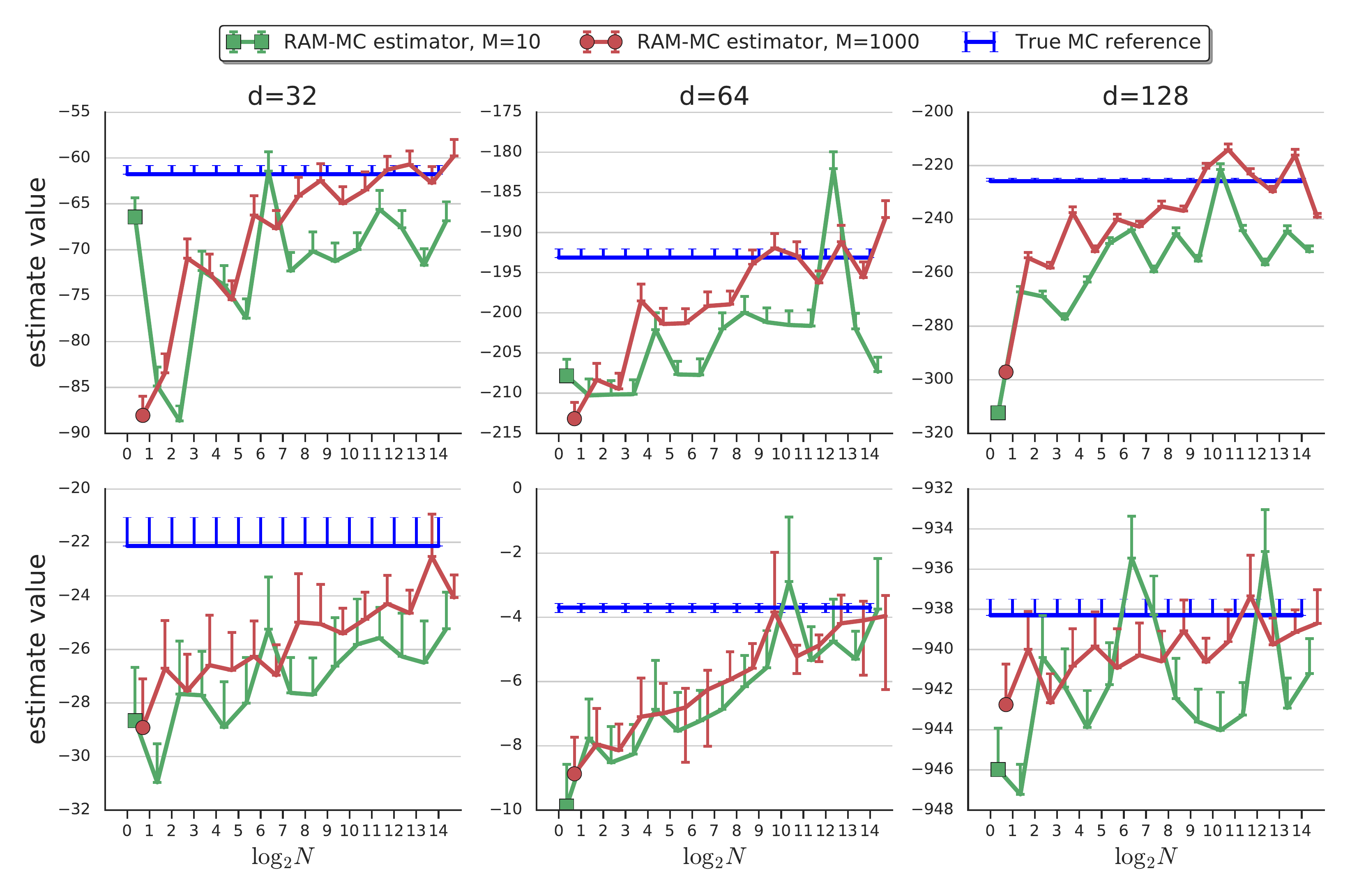}
\end{center}
\caption{\label{fig:real-exps-hsq}
Estimating $H^2(Q_Z^\theta \| P_Z)$ in pretrained autoencoder models with RAM-MC as a function of $N$ for $M=10$ ({\bf \textcolor{green!65!blue}{green}}) and $M{=}1000$ ({\bf \textcolor{red}{red}}) compared to ground truth ({\bf\textcolor{blue}{blue}}).
Lines and error bars represent means and standard deviations over 50 trials.
Plots depict $\log\big(2 - \hat{D}^M_{H^2}(\hat{Q}^N_Z \| P_Z)\big)$ since $H^2$ is close to 2 in all models.
Omitted lower error bars correspond to error bars going to $-\infty$ introduced by $\log$.
Note that the approximately \emph{increasing} behaviour evident here corresponds to the expectation of RAM-MC \emph{decreasing} as a function of $N$. 
Due to concavity of $\log$, the decrease in variance when increasing $M$ manifests itself as the {\bf \textcolor{red}{red}} line ($M{=}1000$) being consistently above the {\bf \textcolor{green!65!blue}{green}} line ($M{=}10$).
}
\end{figure}

Figure~\ref{fig:real-exps-hsq} we display similar results to those displayed in Figure~\ref{fig:real-exps} of the main paper but with the $H^2$-divergence instead of the KL.
An important point is that $H^2(A,B) \in [0, 2]$ for any probability distributions $A$ and $B$,
and due to considerations of scale we plot the estimated values $\log\big(2 - \hat{D}^M_{H^2}(\hat{Q}^N_Z \| P_Z)\big)$.
Decreasing bias in $N$ of RAM-MC therefore manifests itself as the lines \emph{increasing} in Figure~\ref{fig:real-exps-hsq}. 
Concavity of $\log$ means that the reduction in variance when increasing $M$ results in RAM-MC with $M{=}1000$ being above RAM-MC with $M{=}10$.
Similar to those presented in the main part of the paper, these results therefore also support the theoretical findings of our work.

We additionally attempted the same experiment using the $\chi^2$-divergence but encountered numerical issues.
This can be understood as a consequence of the inequality $e^{\KL(A, B)} - 1 \leq \chi^2(A,B)$ for any distributions $A$ and $B$. 
From Figure~\ref{fig:real-exps} we see that the $\KL$-divergence reaches values higher than $1000$ which makes the corresponding value of the $\chi^2$-divergence larger than can be represented using double-precision floats.

\end{document}